\documentclass{article}
\usepackage{ijcai23}

\usepackage{times}
\usepackage{soul}
\usepackage{url}
\usepackage[hidelinks]{hyperref}
\usepackage[utf8]{inputenc}
\usepackage[small]{caption}
\usepackage{graphicx}
\usepackage{amsmath}
\usepackage{amsthm}
\usepackage{amsfonts}
\usepackage{booktabs}
\usepackage{algorithm}
\usepackage{algorithmic}
\usepackage[switch]{lineno}

\usepackage{amsmath,amssymb,amsfonts}
\usepackage{graphicx}
\usepackage{textcomp}
\usepackage{xcolor}
\usepackage{url}
\usepackage{listings}
\usepackage{helvet}
\usepackage{courier}
\usepackage{tabularx}
\usepackage{longtable}
\usepackage{caption}
\usepackage{subfigure}
\usepackage{xspace} 
\usepackage{upgreek}
\usepackage{amsthm}
\usepackage{pgfplots}
\pgfplotsset{compat=1.17} 
\usepackage{scalefnt} 
\newcommand{\eat}[1]{}

\newcommand{\laks}[1]{}

\newtheorem{theorem}{Theorem}
\newtheorem{lemma}{Lemma}

\definecolor{color1}{RGB}{153,21,43}
\definecolor{color2}{RGB}{34,122,68}
\definecolor{color3}{RGB}{32,99,23}
\definecolor{color4}{RGB}{23,89,124}
\definecolor{darkred}{rgb}{0.8, 0.0, 0.0}
\definecolor{darkblue}{rgb}{0.0, 0.0, 0.85}

\newcommand{\negative}[1]{\textcolor{darkred}{#1}}
\newcommand{\positive}[1]{\textcolor{darkblue}{#1}}

\title{KEST: Kernel Distance Based Efficient Self-Training for Improving \\Controllable Text Generation}

\author{
Yuxi Feng$^1$
\and
Xiaoyuan Yi$^2$\footnote{Work done during Yuxi Feng's internship at Microsoft Research Asia mentored by Xiaoyuan Yi.}\and
Laks V.S. Lakshmanan$^1$\And
Xing Xie$^2$
\affiliations
$^1$The University of British Columbia, Vancouver, Canada\\
$^2$Microsoft Research Asia, Beijing, China\\
\emails
\{fyx14,laks\}@cs.ubc.ca,
\{xiaoyuanyi, xing.xie\}@microsoft.com,
}

\begin{document}

\maketitle

\begin{abstract}
Self-training (ST) has come to fruition in language understanding tasks by producing pseudo labels, which reduces the labeling bottleneck of language model fine-tuning. Nevertheless, in facilitating semi-supervised controllable language generation, ST faces two key challenges. First, augmented by self-generated pseudo text, generation models tend to over-exploit the previously learned text distribution, suffering from mode collapse and poor generation diversity. Second, generating pseudo text in each iteration is time-consuming, severely decelerating the training process. In this work, we propose KEST, a novel and efficient self-training framework to handle these problems. KEST utilizes a kernel-based loss, rather than standard cross entropy, to learn from the soft pseudo text produced by a shared non-autoregressive generator. We demonstrate both theoretically and empirically that KEST can benefit from more diverse pseudo text in an efficient manner, which allows not only refining and exploiting the previously fitted distribution but also enhanced exploration towards a larger potential text space, providing a guarantee of improved performance. Experiments on three controllable generation tasks demonstrate that KEST significantly improves control accuracy while maintaining comparable text fluency and generation diversity against several strong baselines. \footnote{To appear in the proceedings of IJCAI 2023.}

\end{abstract}

\section{Introduction}
\label{sec:introduction}

Recent years have witnessed the excellence of Pretrained Language Models (PLMs)~\cite{Liu2019RoBERTaAR,dong2019unified,radford2019language,2020t5} in Natural Language Processing (NLP). However, these PLMs still rely on increasingly more labeled instances for fine-tuning with growing model size~\cite{yogatama2019learning}, hampering their effectiveness under insufficient data~\cite{zhang2020revisiting}. To solve this problem, a promising approach is \emph{Self-training (ST)}~\cite{scudder1965probability,yarowsky1995unsupervised,grandvalet2004semi}, a classic semi-supervised learning~\cite{chapelle2009semi} paradigm. ST minimizes the prohibitively expensive human labeling by iteratively pseudo-annotating unlabeled data with a classifier which is then retrained with the augmented labels. In this way, ST benefits from a vast number of unlabeled instances and extends the generalization bound~\cite{wei2020theoretical,zhang2022does}, boosting a wide spectrum of tasks like Image Classification~\cite{han2019deep,Xie2020SelfTrainingWN}, Speech Recognition~\cite{park2020improved}, and Natural Language Understanding (NLU)~\cite{mukherjee-awadallah-2020-ust,vu-etal-2021-strata,li-etal-2021-task-adaptive}. 

Nonetheless, it is unresolved how to incorporate ST into the data-intensive attribute-controllable Natural Language Generation (NLG), i.e., generate a textual sequence satisfying the input attribute label, as opposed to NLU. Since model inputs now are discrete labels, massive high-quality unlabeled target text (\textit{e.g.}, movie reviews for sentiment-controllable NLG) is essential to construct pseudo label-text pairs, which is impractical in low-resource domains, impeding the broad application of ST~\cite{du-etal-2021-self}. Consequently, classical ST only works for a few generation tasks with adequate plain text, like Sequence Labeling~\cite{wang2020adaptive} and Machine Translation~\cite{He2020Revisiting,jiao-etal-2021-self}.

With limited unlabeled text, a potential approach to further improve ST performance is to leverage the generative ability of NLG models and produce synthetic (pseudo) text~\cite{yang2020generative,schick2021generating} from given labels besides pseudo labels from text. In this case, unfortunately, two major challenges arise. \textbf{i) Over-exploitation}: Augmented by self-generated text, NLG models are forced to repeatedly fit the already learned text distribution. This gradually homogenizes the generated pseudo text and causes a shrunken (collapsed) generalization boundary, resulting in decreased controllability and generation diversity. \textbf{ii) Training deceleration}: We need to re-generate all pseudo text in each ST iteration with updated model parameters, which interrupts the parallelism of Transformer~\cite{vaswani2017attention}-based models, severely decelerating training and impairing practicality.

To tackle these challenges, we propose a novel self-training framework, \textbf{K}ernel Distance Based \textbf{E}fficient \textbf{S}elf \textbf{T}raining (\textbf{KEST}), for improving semi-supervised controllable NLG. Instead of learning from generated pseudo textual sequences with  traditional cross-entropy loss, KEST directly fits the approximated text distribution from the last iteration in the embedding space. Such an objective not only relaxes the constraint imposed by the previous ST iteration but also encourages diverse outputs of the current model, addressing \emph{Challenge (i)}. Besides, we design a non-autoregressive generation schema to produce soft representations of pseudo text (rather than hard strings) in parallel, greatly reducing time cost and handling \emph{Challenge (ii)}. Furthermore, such a soft text is naturally a kind of noisy pseudo data~\cite{He2020Revisiting,Xie2020SelfTrainingWN}, which helps the model denoise errors and propagate local smoothness~\cite{wei2020theoretical,chen-etal-2021-revisiting}. Our method can be theoretically interpreted as exploring a larger potential text space, leading to an extended generalization boundary and improved controllability while maintaining comparable generation quality. \footnote{Code available at \url{https://github.com/peterfengyx/KEST}.}

In summary, our contributions are as follows:
\begin{itemize}
\item We dig into the over-exploitation problem of applying self-training to controllable NLG and propose a novel kernel-based ST framework to address this problem.

\item We design a non-autoregressive generation schema to reduce the time cost of producing pseudo text for self-training, making ST more practical for real scenarios.

\item We theoretically show that KEST could explore a larger potential text space and demonstrate through exhaustive experiments that our model significantly improves controllability with competitive generation diversity and quality, further exploring the capacity frontier of PLMs.

\end{itemize}
\section{Related Work}
\paragraph{Controllable Natural Language Generation}
\label{sec:relatedwork}
Attribute-controllable NLG seeks to make the generated text satisfy user-specified attributes \textit{e.g.}, sentiment, topic and style~\cite{Keskar2019CTRLAC,Dathathri2020Plug} while keeping satisfactory generation quality, which could benefit various downstream applications. With the impressive generation ability of PLMs, a common practice for controllable NLG is fine-tuning a PLM conditioned on attribute labels with attribute-text paris~\cite{Keskar2019CTRLAC,gururangan-etal-2020-dont}. However, as the scale of PLMs keeps increasing, insufficient labeled data becomes a new obstacle to fine-tuning~\cite{yogatama2019learning,zhang2020revisiting}.

To alleviate this problem, another line of methods, called \emph{Plug-in Control}, has been established, which manipulates the output generation probability of models to encourage attribute-related tokens. The manipulation is achieved broadly through two paradigms: updating cached hidden states~\cite{Dathathri2020Plug} or reshaping the output distribution guided by off-the-shelf attribute classifiers~\cite{krause2021gedi,yang2023uddia} or conditional PLMs~\cite{liu-etal-2021-dexperts} at  inference time without fine-tuning. Despite reduced labeling costs, with weak/sparse attribute signals, these methods usually hurt control accuracy or generation fluency.

\paragraph{Self-training} Self-training (ST)~\cite{yarowsky1995unsupervised,grandvalet2004semi} has recently found renewed interest and exhibited notable advantages of augmenting PLM fine-tuning. This paradigm iteratively produces pseudo labels for massive unlabeled data and reduces labeling bottleneck, facilitating varied downstream tasks where massive unlabeled in-domain text exists, including NLU~\cite{vu-etal-2021-strata,du-etal-2021-self,bhat-etal-2021-self,chen-etal-2021-revisiting}, Image Classification~\cite{han2019deep,Xie2020SelfTrainingWN,sohn2020fixmatch}, Speech Recognition~\cite{park2020improved,kahn2020self}, and Neural Machine Translation (NMT)~\cite{zhang2016exploiting,He2020Revisiting,jiao-etal-2021-self}. 

Besides classical ST, diverse follow-up modifications have been developed for further improvement, which generally fall into two lines. The first line, \textit{i.e.}, sample selection, selects only a part of unlabeled instances in terms of (1) model confidence to avoid over-noisy pseudo labels~\cite{sohn2020fixmatch,bhat-etal-2021-self}, (2) prediction uncertainty to obtain informative instances and enhance performance on the hard ones~\cite{mukherjee-awadallah-2020-ust,jiao-etal-2021-self}, or (3) label balance to benefit minority classes~\cite{wei2021crest}. The other line is noisy labeling~\cite{He2020Revisiting,Xie2020SelfTrainingWN}, which injects synthetic noise into the pseudo data, \textit{e.g.}, token shuffle or image distortion to propagate local smoothness and improve model robustness.

Despite remarkable progress, as discussed in Sec.\ref{sec:introduction}, these ST methods are unsuitable for attribute-controllable NLG because of the two challenges identified earlier. 

\paragraph{Non-Autoregressive Generation (NAG)} Relevant to our work, NAG aims to simultaneously generate all target tokens rather than one by one to increase the inference speed. NAG was first proposed in NMT~\cite{gu2018non,ma2019flowseq} and then applied to broader scenarios like Text Summarization~\cite{liu2022learning} and Text-to-Speech Synthesis~\cite{chien2021hierarchical}. All the tasks are learned with encoder-decoder architectures, relying on long input sequences (\textit{e.g.}, source language) to provide rich initial context information. However, it is still challenging to leverage NAG for our task since the inputs are only attribute labels and short prompts.

Unlike the work above, we take a further step to investigate the challenges of incorporating ST with controllable NLG and propose a practical NAG method to generate soft pseudo text, which is then learned in a kernel space, leading to a novel and efficient ST framework.

%\paragraph{Kernel-based MMD loss} Maximum Mean Discrepancy (MMD) loss has been proposed to minimize Wasserstein distance in the training of Wasserstein GAN \cite{arjovsky2017wasserstein} and Wasserstein Auto Encoders \cite{tolstikhin2018wasserstein} which are widely used in image generation.   
\section{Method}

\subsection{Formulation and Overview}
Let $\mathbf{x}_i$ denote a textual sequence and $y_i$  an attribute label. Assume we have a labeled dataset $D_l\!=\!\{\mathbf{x}_i,y_i\}_{i=1}^{N_l}$, and an unlabeled in-domain set $D_u\!=\!\{\mathbf{x}_i\}_{i=1}^{N_u}$ where $N_u \gg N_l$. Our goal is to learn an attribute-controllable generator $\mathcal{G}_{ag}(y)\!=\!P_{\theta}(\mathbf{x}|y)$ (parameterized by $\theta$) to generate high-quality text $\mathbf{x}$, matching the given label $y$. In addition, we endow the generator with the ability of multi-task generation. Concretely, the model is reused and jointly trained to generate (a) pseudo text $\hat{\mathbf{x}}$ in a non-autoregressive manner, depicted as $\mathcal{G}_{nag}(y)$, for further augmenting self-training, and (b) pseudo labels $\hat{y}$ for $\mathbf{x} \in D_u$, namely, a classifier $\mathcal{C}\!=\!P_{\theta}(y|\mathbf{x})$.

During the self-training phase, besides the pseudo label pairs $(\mathbf{x},\hat{y})$, KEST also learns the pseudo text pairs $(\hat{\mathbf{x}},y)$ in the kernel space to simultaneously cover more unseen instances and extend the previously fitted distribution (Sec.\ref{subsec_theorem}), \emph{handling Challenge (i)}. All the pseudo text $\hat{\mathbf{x}}$ is produced through NAG efficiently, \emph{handling Challenge (ii)}.
%--------------------------------------------------
\subsection{Multi-task Generator}
\label{subsec_dvae}
To further enhance the performance and efficiency of our model, we design a multi-task generator to produce the desired text $\mathbf{x}$, \emph{Pseudo Label (PL)} $\hat{y}$, and \emph{Pseudo Text (PT)} $\hat{\mathbf{x}}$ jointly based on a shared PLM.

\textbf{Autoregressive Text Generation}. To obtain high-quality attribute-specified generated text $\mathbf{x}$, we optimize the generator $\mathcal{G}_{ag}$ in an autoregressive manner as follows:
\begin{align}
    \mathcal{L}_{ag} & = - \frac{1}{N} \sum_{(\mathbf{x},y) \in D} [\sum_{j=1}^L \log P_{\theta}(\mathbf{x}^j|\mathbf{x}^{<j},y)], \label{agloss}
\end{align}
where $\mathbf{x}^j$ means the $j$-th token in $\mathbf{x}$, $L$ is the length of $\mathbf{x}$, $D$ is the training set with $N$ samples. We will show later how to construct $D$ for different training phases. 

\textbf{Pseudo Label Generation}. We also make our model simultaneously learn a classifier $\mathcal{C}$ by minimizing:
\begin{align}
    \mathcal{L}_c &= - \frac{1}{N} \sum_{(\mathbf{x},y) \in D} \log P_{\theta}(y|\mathbf{x}). \label{clloss}
\end{align}

Eq. (\ref{clloss}) enables our model to make full use of available unlabled text $\mathbf{x} \in D_u$ to produce pseudo labels by $\hat{y}=\text{MLP}\left(\text{Encoder}(\mathbf{x})\right)$, helping regularize the training and improve the generalization bound~\cite{wei2020theoretical}. 

\textbf{Non-autoregressive Pseudo Text Generation}. With insufficient unlabeled text, we could produce pseudo text for further improvement and then speed up the repetitive PT generation via NAG. However, as shown in Sec.~\ref{sec:relatedwork}, an input consisting of just $y$ is too uninformative to guide the generation, hampering convergence and causing extremely noisy PT.

To mitigate this problem, we resort to the Masked Language Model (MLM) ~\cite{devlin2019bert} to train the NAG generator $\mathcal{G}_{nag}$ and conduct generation. Define $\mathbf{m} \sim \mathcal{B}(L, p_m)$ as a mask indicator vector, where $\mathcal{B}$ is the Bernoulli distribution. Given a text $\mathbf{x}$, we replace part of the tokens in it with the $\text{MASK}$ symbol and get the masked one $\mathbf{x}^{\setminus\mathbf{m}}=[\mathbf{x}^1,\cdots,\text{MASK},\cdots,\mathbf{x}^L$], where $\mathbf{x}^j=\text{MASK}$ iff $\mathbf{m}^j=1$. Then we optimize the following loss for NAG:
\begin{align}
    \mathcal{L}_{nag} & \!=\!- \frac{1}{N} \sum_{(\mathbf{x},y) \in D} [\sum_{j=1}^L \mathbb{I}(\mathbf{m}_j\!=\!1) \log P_{\theta}(\mathbf{x}^j|\mathbf{x}^{\setminus\mathbf{m}},y)], \label{nagloss}
\end{align}
where $\mathbb{I}$ is the indicator function and the masking probability $p_m$ can be adjusted as the noise level.

In this way, our model only needs to predict partial tokens according to the rich context $\mathbf{x}^{\setminus\mathbf{m}}$, which is easier to learn, reducing the time complexity of PT generation from $\mathcal{O}(L)$ to $\mathcal{O}(1)$ (see Fig. \ref{fig:dectime}). Besides, the pseudo text $\mathbf{\hat{x}}=\mathcal{G}_{nag}(\mathbf{x}^{\setminus\mathbf{m}},y)$ naturally introduces moderate noise in terms of re-predicted tokens while maintaining satisfactory fluency due to the unaltered high-quality ones. Such a flexible corruption acts as a kind of weak augmentation~\cite{chen-etal-2021-revisiting} which enhances the exploitation and outperforms typical synthetic noise (\textit{e.g.}, token dropout) ~\cite{He2020Revisiting}. 

The final loss is computed as follows:
\begin{equation}
\label{dvaeloss}
\begin{aligned}
    \mathcal{L}=&\lambda_c \mathcal{L}_c+\lambda_{ag} \mathcal{L}_{ag}+\lambda_{nag}\mathcal{L}_{nag}
\end{aligned}
\end{equation}
where $\lambda_c$, $\lambda_{ag}$, and $\lambda_{nag}$ are hyper-parameters.

%--------------------------------------------
\subsection{Kernel-based Learning}
\label{sec:mmd}
As we discussed in Sec.~\ref{sec:introduction}, learning from self-generated pseudo text $\mathbf{\hat{x}}$ with the standard cross-entropy loss forces the current model $P_{\theta}$ to over-exploit and is shackled to the previously learned one $P_{\theta^{'}}$ (Sec.~\ref{subsec_theorem}), resulting in a shrunken generalization boundary and decreased controllability.

To break such constraints, we make the current model $P_{\theta}$ directly fit the previous one $P_{\theta^{'}}$. For this goal, we leverage \emph{Maximum Mean Discrepancy (MMD)}~\cite{gretton2012kernel}, a well-known kernel-based probability measure, and minimize the following empirical loss for all generated pseudo text:
\begin{equation}
\label{mmdloss}
\begin{aligned}
\mathcal{L}_{ker} =& \frac{1}{N(N-1)} \sum_{\mathbf{\tilde{x}}_i, \mathbf{\tilde{x}}_j \in D_o, i\ne j} k(\mathbf{\tilde{x}}_i, \mathbf{\tilde{x}}_j)\\
-&\frac{2}{N^2} \sum_{\mathbf{\tilde{x}}_i \in D_o, \mathbf{\hat{x}}_j \in D_{pt}} k(\mathbf{\tilde{x}}_i, \mathbf{\hat{x}}_j),
\end{aligned}
\end{equation}
where $D_{pt}=\{ \mathbf{\hat{x}}_i \}_{i=1}^{N}$ is set of pseudo text, $D_{o}=\{ \mathbf{\tilde{x}}_i \}_{i=1}^{N}$ is set of text generated by $\mathcal{G}_{ag}(\mathbf{\hat{x}}_i,y)$ (or $\mathcal{G}_{nag}(\mathbf{\hat{x}}_i^{\setminus\mathbf{m}},y)$) in the self-training phase. $k$ is the kernel function, for  which we take the RBF kernel here, that is, $k(\mathbf{\tilde{x}}_i, \mathbf{\tilde{x}}_j)=\text{exp}\left(\frac{-\|\mathbf{\tilde{x}}_i-\mathbf{\tilde{x}}_j\|^2}{2\sigma^2}\right)$ and $\sigma$ is the bandwidth.

This MMD loss is an unbiased estimator and  model parameters can be learned through back-propagation. We will demonstrate in Sec.~\ref{subsec_theorem} that such an objective could relax the constraint imposed by the previous model $P_{\theta^{'}}$  and encourage more diverse outputs.

%-----------------------------------
\begin{algorithm}[tp]
    \caption{Training Process of KEST}
    \label{alg:kest}
    \textbf{Input}: Labeled set $D_l$, unlabeled set $D_u$\\
    \textbf{Output}: The trained model $P_{\theta}$
    \begin{algorithmic}[1] %[1] enables line numbers
        \STATE Jointly train base model $\mathcal{G}_{ag}$, $\mathcal{G}_{nag}$ , $\mathcal{C}$ on $D_l$ by optimizing Eq.\eqref{dvaeloss}, store the best $\mathcal{G}_{ag}^0$, $\mathcal{G}_{nag}^0$, $\mathcal{C}^0$.
        \FOR {$epoch\leftarrow 1$ to $MaxEpoch$}
        \FOR{$\mathbf{x}_i$ \textbf{in} $D_u$}
        \STATE $\hat{y}_i=\mathcal{C}^{epoch-1}(\mathbf{x}_i)$
        \ENDFOR
        \STATE Build pseudo labeled dataset $D_{pl} = \{\mathbf{x}_i,\hat{y}_i\}$
        \STATE Sample a subset $D_{pseudo}$ from $D_l\cup D_{pl}$ 
        \FOR {$(\mathbf{x}_i,y_i)$ \textbf{in} $D_{pseudo}$}
        \STATE Sample mask vector $\mathbf{m}$.
        \STATE $\mathbf{\hat{x}}_i=\mathcal{G}_{nag}^{epoch-1}(\mathbf{x}^{\setminus\mathbf{m}}_i,y_i)$
        \ENDFOR

Build pseudo text dataset: $D_{pt}=\{\mathbf{\hat{x}}_i,y_i\}$

        \STATE Train $\mathcal{G}_{ag}^{epoch-1}$, $\mathcal{G}_{nag}^{epoch-1}$, and $\mathcal{C}_{epoch-1}$ on $\{D_{pt},D_{pl}, D_l\}$ by optimizing Eq.\eqref{dvaeloss} and Eq.\eqref{sptloss}, update the parameter to $\mathcal{G}_{ag}^{epoch}$, $\mathcal{G}_{ag}^{epoch}$, and $\mathcal{C}^{epoch}$.
        \ENDFOR
    \end{algorithmic}
\end{algorithm}
%------------------------------------------

\paragraph{Soft Pseudo Text (SPT)} When optimizing Eq.~(\ref{mmdloss}), we need to calculate the $l_2$-distance between two text $\mathbf{x}_i$ and $\mathbf{x}_j$. Simply using hard text (one-hot representations) has two drawbacks. First, the signal would be too sparse since most dimensions are zeros in the vector. Second, the sampled discrete $\mathbf{x}_i$ (a point in the text space) causes information loss and forces us to sample numerous points to cover a small neighborhood region in the space.

Therefore, we further propose to generate soft pseudo text. We use the feature representation of the text $\mathbf{x}$, $e(\mathbf{x})= P(\mathbf{x}) \! \times \! \mathbf{E} \! \in \! \mathbb{R}^{L\times d}$, where $P(\mathbf{x}) \! \in \! \mathbb{R}^{L\times V}$ are the generation probabilities of each token $\mathbf{x}^i$ on the vocabulary, and $\mathbf{E} \! \in \! \mathbb{R}^{V\times d}$ is the word embedding matrix. $V$ and $d$ are vocabulary and embedding sizes, respectively. Then we change Eq.(\ref{agloss}) and Eq.~(\ref{nagloss}) to:

\begin{equation}
\begin{aligned}
 \mathcal{L}_{ag}^{'} & = \mathcal{L}_{ker} \ \text{if} \  \mathbf{x} \in D_{pt} \  \text{else} \  \mathcal{L}_{ag} \\
 \mathcal{L}_{nag}^{'} & = \mathcal{L}_{ker} \ \text{if} \  \mathbf{x} \in D_{pt} \  \text{else} \  \mathcal{L}_{nag}.
\end{aligned}
\label{sptloss}
\end{equation}

In this way, we avoid losing relevant semantics information in the pseudo text, make the model fit a smoother distribution and further extend the generalization boundary (see Table~\ref{tab:ablation}).

Following the practice of self-training in NLU~\cite{vu-etal-2021-strata}, we start ST from a strong base model tuned on $D_l$ and use the full unlabeled $D_u$ to produce pseudo labels, rather than select part of the data with certain criteria as in~\cite{mukherjee-awadallah-2020-ust,jiao-etal-2021-self}. The PLM word embedding $\mathbf{E}$ is frozen during self-training. The complete KEST process is described in Alg.~\ref{alg:kest}.

%------------------------------------------
\subsection{Further Analysis of KEST}
\label{subsec_theorem}
%-------------------------------------
\begin{figure}[htp]
\center
\includegraphics[width=0.46\textwidth]{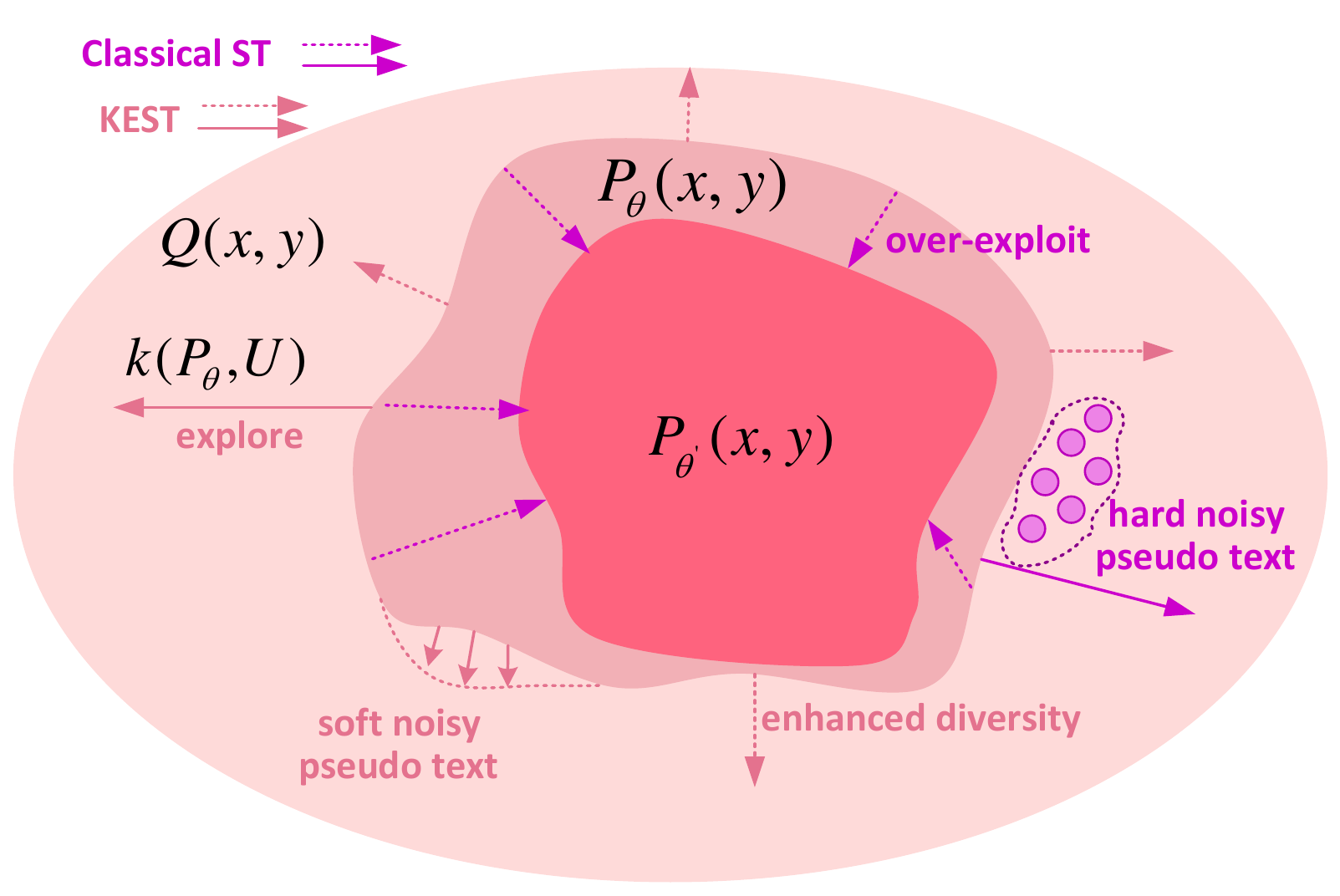}
\caption{The illustration of KEST advantages.} 
\label{fig_theorem} 
\end{figure}
%--------------------------------------
To better understand the advantages of KEST, we provide the following two results.
\begin{lemma}
\label{lemma}
The optimization of classical self-training is equivalent to minimizing $(1-\alpha)*KL[Q(x,y)||P_{\theta}(x,y)]+\alpha*KL[P_{\theta^{'}}(x,y)||P_{\theta}(x,y)]$, where $Q$ is the real joint distribution of text and label, $P_\theta$ and $P_{\theta'}$ are models estimated at the current and last ST iteration, respectively, $KL$ is the Kullback–Leibler divergence, and $\alpha$ is the ratio of pseudo text.
\end{lemma}

From Lemma~\ref{lemma}, we can see that classical ST approximates the text distribution and fits the current model into the previously learned one. Since $\text{KL}[P_{\theta^{'}}(x,y)||P_{\theta}(x,y)]=\iint  P_{\theta^{'}}(x,y) \log \frac{P_{\theta^{'}}(x,y)}{P_{\theta}(x,y)} dx dy$, failing to assign enough probability mass to a point $(x,y)$ in $P_{\theta^{'}}$ will bring extremely larger loss. Consequently, $P_{\theta}$ is more inclined to cover $P_{\theta^{'}}$ rather than explore $Q$, causing over-exploitation.

In contrast, we give a theorem of our KEST:
\begin{theorem}
\label{thrm}
Minimizing the training objective of KEST is equivalent to minimizing the following: 
\begin{equation}
\label{eqkl}
\begin{aligned}
    &\text{KL}\left[Q(x,y)||P_\theta(x,y)\right] \\
    + & \text{MMD}^2 \left[P_{\theta'} (x,y)||P_\theta (x,y)\right] \\
    - & 2* \mathbb{E}_{P_{\theta}U} \left[ k(x,u)\right],
\end{aligned}
\end{equation}
where $U$ is a noise distribution.
\end{theorem}
\emph{Proof}. See Appendix~\ref{sec:apdix-theorem}.

In Theorem \ref{thrm}, our KEST fits the true distribution $Q$ by KL divergence to cover the real space as large as possible while fitting $P_{\theta^{'}}$ with MMD. Considering Eq.(\ref{mmdloss}), we can see this loss not only regularizes $P_{\theta}$ by $P_{\theta^{'}}$, but also diversifies $P_{\theta}$ via increasing the $l_2$-distance of generated text $\|\mathbf{x}_i-\mathbf{x}_j\|^2$, and enhances exploration through fitting a noise distribution and disturbing $P_{\theta}$, further pushing the generalization boundary.
\begin{table*}[htbp]
%\small
 \centering
 \begin{tabular}{lcccccccccl}\toprule
    & \multicolumn{5}{c}{Sentiment} & \multicolumn{5}{c}{Topic}
    \\
    \cmidrule(lr){2-6}\cmidrule(lr){7-11}
             & O-PPL $\downarrow$  & M-PPL $\downarrow$ & F1 $\uparrow$ & Dist $\uparrow$ &S-BLEU $\downarrow$  & O-PPL $\downarrow$  & M-PPL $\downarrow$ & F1 $\uparrow$ & Dist $\uparrow$ &S-BLEU $\downarrow$ \\\midrule
    Test set & 25.14&$-$ & 96.20 & 48.27 & 43.34 & 31.04&$-$ & 94.89 & 67.24 & 23.31 \\
    GPT2 (raw) & 13.20&38.39&68.50&35.91&58.79 & 16.94 & 74.41 & 52.17 & 46.88&45.55\\
    \midrule
    \multicolumn{7}{l}{\textit{Fine-tuned PLM}} \\
    GPT2    &\textbf{16.40}&44.02&80.44&26.34&71.00 &\textbf{22.22 }&23.46&83.08&54.93&\underline{39.93}\\
    UniLM & 25.20&54.33&75.35&31.05&66.97&55.79 &36.28&87.70&54.76&43.77 \\
    T5 & 25.69&34.97&83.77&30.03&69.57
    & 48.33&32.12&88.43&\textbf{58.06}&\textbf{37.01}\\
    %\midrule
    %\multicolumn{7}{l}{\textit{Lightweight method}}\\
    %PF  &13.02&37.09&75.05&29.48&65.10
    %&\textbf{20.27}&32.35&68.44&59.17&32.73\\
    %Ctr-PF &\textbf{13.01}&37.12&77.33&29.63&\textbf{64.83}
    %&20.41 &33.90&83.21&\textbf{60.34}&\textbf{31.20}\\
    \midrule
    \multicolumn{7}{l}{\textit{Self-Training methods}}\\
    PT    & 26.62 &58.37&70.27&31.17&66.69 & 57.40&40.95&86.36&52.35&46.41\\
    PT(noise) &30.28&62.07&75.78&\underline{31.68}&\textbf{65.18}& 58.59&45.32&85.27&53.35&46.57 \\
    PT(noise)+PL &18.92&\textbf{33.53}&89.73&30.94&66.84&32.36&\textbf{16.64}&89.70&53.79&47.95 \\
    PT(select)+PL &\underline{18.40}&\underline{33.56}&\underline{90.06}&31.27&67.61&33.23&\underline{16.66}&\underline{90.52}&53.71&47.69 \\
    \midrule
    %\multicolumn{7}{l}{\textit{Our Methods}}\\
    KEST  & 20.65 & 38.15 & \textbf{91.77} & \textbf{31.70} & \underline{66.60} & \underline{31.19} &20.46 &\textbf{91.94}&\underline{56.16}&42.10\\
    \bottomrule
 \end{tabular}
 \caption{Automatic evaluation results on IMDb dataset (sentiment) and AGNews dataset (topic). For each metric, the best results are in \textbf{bold}, and the second-best results are \underline{underlined}.}
 \label{tab:mainresult}
\end{table*}
\section{Experiments}
\label{sec:experiment}
\subsection{Tasks}
We conduct exhaustive experiments on three controllable generation tasks, described below: 

\textbf{Sentiment control with prompt:} We evaluate the sentiment controllability on the IMDb movie review dataset~\cite{maas-etal-2011-learning}. Following~\cite{Dathathri2020Plug}, we use their 15 prompts and another 85 prompts sampled from IMDb (100 in total) as model input. We generate 10 samples for each prompt and each sentiment.

\textbf{Topic control w/o prompt:} We use the AGNews dataset ~\cite{Zhang2015CharacterlevelCN} to evaluate topic controllability. We assess our model's ability to generate from scratch on this dataset and sample 300 generations for each topic.

\textbf{Text detoxification:} We use the Jigsaw Toxicity Dataset for training, and use the 203 ``challenging'' prompts (toxicity$\!<\!0.5$) from~\cite{gehman-etal-2020-realtoxicityprompts} to generate 10 non-toxic sentences for each prompt, following~\cite{qian-etal-2022-controllable}.

For IMDb, we sample 5\% of the training samples as labeled data and directly take their provided unlabeled set. Since there is no separate unlabeled text in AGNews, we sample 3\% of training samples as labeled data and use the others as unlabeled ones. For a fair comparison, we keep the ratio of labeled/pseudo/unlabeled text to 1:1:30. More details of the dataset are provided in Appendix \ref{sec:datasetdesc}.
%----------------------------------------------
\subsection{Experimental Settings}
We use UniLM-base-cased \cite{dong2019unified} as the shared classifier and generator. We use AdamW~\cite{Loshchilov2019DecoupledWD} with learning rate $=$ 5e-5, warm-up steps $=$ one epoch, and batch size $=$ 8 for optimization. The top-$p$ ($p=0.9$) sampling method is used for decoding in evaluation. We set $\lambda_c\!=\!\lambda_{ag}\!=\!\lambda_{nag}\!=\!1.0$ in Eq. (\ref{dvaeloss}) across all tasks. More implementation details are provided in Appendix \ref{sec:impdetail}.

\subsection{Evaluation Metrics}
We mainly focus on improving control accuracy and diversity while maintaining the generation quality in this work, considering the following four kinds of metrics. We also provide the classification performance of KEST in Appendix \ref{sec:apdix-toxic}.

\textbf{Fluency:} We evaluate generation fluency by the perplexity of generated text measured by a GPT2-XL~\cite{radford2019language} model, \textit{i.e.}, \textbf{Output PPL}.

\textbf{Generalizability:} We calculate the perplexity of each model on each held-out test set provided in each dataset, \textit{i.e.}, \textbf{Model PPL}, which measures how well the model generalizes and adapts to the unseen domain under a specified attribute.

\textbf{Controllability:} We evaluate the control accuracy through classification Macro-F1 (\textbf{F1}) on the generated text by two RoBERTa-large classifiers fine-tuned on the full IMDb and AGNews data (testing F1=96\% and 95\%), respectively. For toxicity evaluation, we use the Perspective API\footnote{\url{https://www.perspectiveapi.com/}}.

\textbf{Diversity:} To evaluate the diversity of generated text, we consider \textbf{Dist-n}~\cite{li-etal-2016-diversity} and \textbf{Self-BLEU}~\cite{zhu2018texygen}. More metrics details are described in Appendix \ref{sec:apdix-metric}.

\subsection{Baselines}
We compare our model with the following (supervised or semi-supervised) strong NLG baselines.

\textbf{Fine-tuned PLM}: We fine-tune diverse powerful PLMs on each of the datasets, including GPT2~\cite{radford2019language}, UniLM~\cite{dong2019unified} and T5~\cite{2020t5}.

\textbf{Self-training methods:}\\
(1) PT: the naive Self-training~\cite{grandvalet2004semi}, which generates pseudo text at each epoch and updates parameters with both real and pseudo text.\\
(2) PT(noise): the noisy version of Self-training~\cite{He2020Revisiting}, which brings synthetic noise (token drop, swap and mask) to the pseudo text for self-training.\\
(3) PT(noise)+PL: We combine PT(noise) with \emph{pseudo labeling}, and fine-tune a BERT-base~\cite{devlin2019bert} on $D_l$ 
to generate pseudo labels for all real unlabeled text.\\
(4) PT(select)+PL: The sample selection version of ST~\cite{mukherjee-awadallah-2020-ust}. We over-generate noisy pseudo text and select samples by the classifier confidence and uncertainty scores.

All the self-training methods are applied to the same fine-tuned UniLM as used in KEST. We give more details of the baseline models above in Appendix \ref{sec:apdix-baseline}.

\subsection{Results}
As shown in Table \ref{tab:mainresult}, in both sentiment (with prompt) and topic (without prompt) controlled generation, our KEST achieves significant improvement in control accuracy (+8.0 F1 at most) compared to fine-tuned PLMs. The generally much higher PPL (for UniLM and T5), limited F1 improvement, and severely decreased diversity (for GPT2) indicate these PLMs either fail to be adapted to new domains (\textit{e.g.}, positive movie reviews) or overfit with inadequate labeled data, as analyzed in~\cite{zhang2020revisiting}. On the contrary, thanks to the self-augmented data, KEST notably enhances controllability as well as fluency and diversity, especially compared to the backbone UniLM.

We also observed some interesting results considering existing self-training methods. 1) The naive self-training with PT performed poorly in controllability and diversity, even worse than tuned PLMs, due to the over-exploitation and shrunken distributions as interpreted in Sec.~\ref{subsec_theorem}. 2) The traditional synthetic noise (PT(noise)) slightly boosts control accuracy and diversity, which verifies the effectiveness of noise~\cite{He2020Revisiting} again, but greatly hurts fluency and generalizability (+3.7 O-PPL at most). This is because  such hard corruption is too noisy and makes the model diverge far from valid attribute distributions. In contrast, KEST utilizes a NAG generator to produce flexible noise, improving local smoothness. 3) Additional pseudo-labels bring significant improvement, especially on PPL. However, with a fixed number of unlabeled data, the performance of these methods is limited. Besides, KEST utilizes the multi-task generator to produce soft pseudo text in feature space, which helps cover a larger attribute space and obtain further improvement.

Due to space limitations, we report the results and experiment details of text detoxification under both automatic and human evaluation in Appendix \ref{sec:apdix-toxic}.

\subsection{Human Evaluation}
\begin{table}[thp]
\small
 \centering
 \begin{tabular}{rlll}\toprule
         & Fluency $\uparrow$  & Novelty $\uparrow$ & Rel. $\uparrow$\\\midrule
    \textit{Sentiment} &\\
    UniLM-PT(select)+PL & 3.60 &3.40 & 3.62$^{**}$\\
    KEST  & \textbf{3.67} & \textbf{3.48} & \textbf{3.87} \\
    \midrule
    \textit{Topic} \\
    UniLM-PT(select)+PL &3.97$^{**}$ &4.43 & 4.57\\
    KEST  & \textbf{4.11}&\textbf{4.54}&\textbf{4.62}\\
    \bottomrule
 \end{tabular}
 \caption{Human evaluation results on sentiment/topic-controlled generation.  We conduct the Student t-test to evaluate statistical significance ($^{**}$: $p$-value$<0.01$). The overall Cohen's kappa score is 0.62, showing a satisfactory inter-annotator agreement.}
 \label{tab:human}
\end{table}
To better verify the effectiveness of KEST, we also conduct a human evaluation. For each model, we generated 100 samples on each task. We invite 6 competent annotators to score these samples on three criteria --  \textbf{Fluency}, \textbf{Novelty}, and \textbf{Attribute Relevance} in a blind review manner. As shown in Table \ref{tab:human}, KEST consistently outperforms the best baseline (UniLM-PT(select)+PL) on all three metrics, which indicates that KEST not only has better controllability over attributes but also generates fluent and diverse texts. See Appendix \ref{sec:apdix-human} for detailed evaluation descriptions and metrics.

\subsection{Ablation Study}
\begin{table}[htbp]
\small
 \centering
 \begin{tabular}{rcccl}\toprule
    & \multicolumn{4}{c}{AGNews} %& \multicolumn{3}{c}{Jigsaw}
    \\
    \cmidrule(lr){2-5}%\cmidrule(lr){5-7}
             & O-PPL $\downarrow$  &M-PPL $\downarrow$ & F1 $\uparrow$ &S-BLEU $\downarrow$  \\\midrule
    KEST  & \textbf{31.19} &\textbf{20.46} &\textbf{91.94}&\textbf{42.10}\\
    $-$Soft & 38.04&29.07 &90.96&44.09\\
    $-\mathcal{L}_{ker}-$Soft &38.98 &28.77 &90.81&45.02\\
    $\!-\!\mathcal{L}_{nag}\!-\!\mathcal{L}_{ker}-$Soft & 39.73 &28.58&90.42&44.73\\
    $-$PT&38.09&28.77&90.97&44.13\\
    $-$PL$-$PT&37.24&256.66&87.45&69.30\\
    \bottomrule
 \end{tabular}
 \caption{Ablation study on AGNews dataset. The symbol $-$ means removing the settings from KEST. $-$Soft: using sampled hard tokens instead of the soft $e(\mathbf{x})$. $-\mathcal{L}_{ker}$: using the cross-entropy loss instead of Eq.(\ref{mmdloss}). $-\mathcal{L}_{nag}$: using $\mathcal{G}_{ag}$ to generate pseudo text instead of $\mathcal{G}_{nag}$. $-$PT/$-$PL: do not use pseudo text/labels.}
 \label{tab:ablation}
\end{table}
We conduct an ablation study on the AGNews dataset and compare different KEST variants. As shown in Table \ref{tab:ablation}, we can find: 1) Soft pseudo text prominently improves PPL and diversity, outperforming the hard one. As discussed in Sec.~\ref{sec:mmd}, such soft PT could bring smoother noise and help further push the learned distribution boundary. 2) Kernel-based learning alleviates the over-exploitation problem of the traditional cross-entropy loss and further enhances inner-group diversity (-0.93 S-BLEU), empirically supporting Theorem~\ref{thrm}. 3) KEST's NAG ability not only reduces time complexity but also slightly boosts fluency (-0.75 O-PPL). However, the diversity improvement attributed to NAG is correlated to the noisy level. Only with an appropriate masking probability $p_m$ could NAG facilitate more diverse text (see Fig.~\ref{fig:mask}). Besides, pseudo text notably promotes all metrics, verifying our claim in Sec.~\ref{sec:introduction} that such synthetic pseudo text leads to further improvement beyond pseudo labels.  

\subsection{Analysis}
\label{sec:analysis}
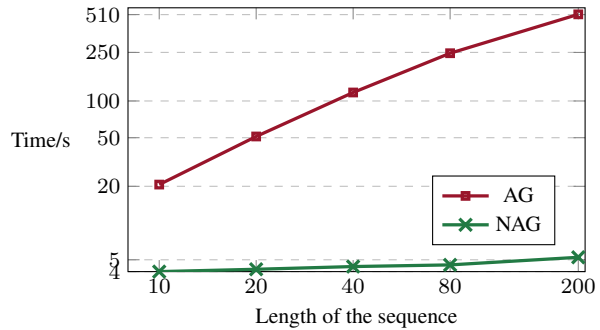
\begin{figure}[thp] %插入图片
\centering %图片居中
\resizebox{0.95\columnwidth}{!}{  %用于修改图片大小
			\begin{tikzpicture} %tikz图片
			\scalefont{0.8} %设置字体大小
			\begin{axis}[
			sharp plot, %控制线的风格
			%title=Generation Control,%图像标题
			xmode=log,% 控制坐标轴为线性
    		ymode=log,% 控制坐标轴为对数
            log ticks with fixed point,
			xlabel=Length of the sequence, %x坐标名
			ylabel=Time/s, %y坐标名
			width=7.5cm, height=5cm,  %设置长和宽
			xmin=8,xmax=210,  % 设置x坐标范围
			ymin=4, ymax=580,  % 设置y坐标范围
			xtick={0,10,20, 40,80,200}, %指定x轴刻度值。如果为空，则自动设置刻度线。即分割坐标轴
			ytick={4,5,20,50,100,250,510}, %指定y轴刻度值。如果为空，则自动设置刻度线。即分割坐标轴
			xlabel near ticks, % 设置x坐标名位置靠近折线图
			ylabel near ticks, % 设置y坐标名位置靠近折线图
                ylabel style={rotate=-90},
			ymajorgrids=true, % 启用/禁用 [公式] 轴上刻度线位置上的网格线
			grid style=dashed, % 设置网格线格式
			legend style={at={(0.8,0.1)},anchor=south}, % 设置标签位置
%			legend columns=3, %设置标签列数
%			legend pos=north west, % 设置折线对应标签的位置
%			legend style={nodes={scale=0.6, transform shape}},  % 设置折线标签的格式
			]
			\addplot+[very thick,mark=square,mark options={scale=0.6}, color=color1] plot coordinates {
                    (10,20.66)
                    (20,51.18)
                    (40,117.36)
                    (80,246.16)
                    (200,512.52)

			};
            \addlegendentry{AG} 

            \addplot+[very thick,mark=x,mark options={scale=1.6}, color=color2] plot coordinates {
                    (10,4.00)
                    (20,4.18)
                    (40,4.40)
                    (80,4.54)
                    (200, 5.24)
			};
            \addlegendentry{NAG}
			\end{axis}

\end{tikzpicture}
		}

		\caption{Comparison of decoding time of NAG and AG for 100 pseudo text batches (batch size=8) with different text lengths.} % 设置caption
		\label{fig:dectime}  % 设置用于reference的label
\end{figure}

\eat{
\begin{table}[htbp]
 \centering
 \begin{tabular}{rcl}
 \toprule
             & IMDb & AGNews \\
             \midrule
    %UniLM+PT(select)+PL & 16h  &14h\\
    KEST(AG) &  21h & 12h  \\
    KEST(NAG) & 17.6h  & 9h\\
    \bottomrule
 \end{tabular}
 \caption{Time consumption of training model for 20 epochs.}
 \label{tab:training_time}
\end{table}}

\textbf{Time Consumption:}
Fig.~\ref{fig:dectime} shows the decoding time of our NAG and AG generators for generating pseudo text with different text lengths. We found that the time costs of the AG module  increase almost linearly w.r.t. the text length. In comparison, our NAG generator $\mathcal{G}_{nag}$ greatly accelerates the generation of pseudo text, especially when the sequence length is long. Furthermore, we compare the training time of KEST using $\mathcal{G}_{ag}$ and $\mathcal{G}_{nag}$, respectively. We observe that the latter achieves \emph{1.2$\times$} and \emph{1.3$\times$} speedup on IMDb and AGNews, respectively, which could be further improved with a larger ratio of pseudo text (e.g., the ratio could increase to \emph{6.7$\times$} with 1/10 unlabeled data), making self-training more practical.

\begin{figure}[ht]
\centering
\includegraphics[scale=0.5]{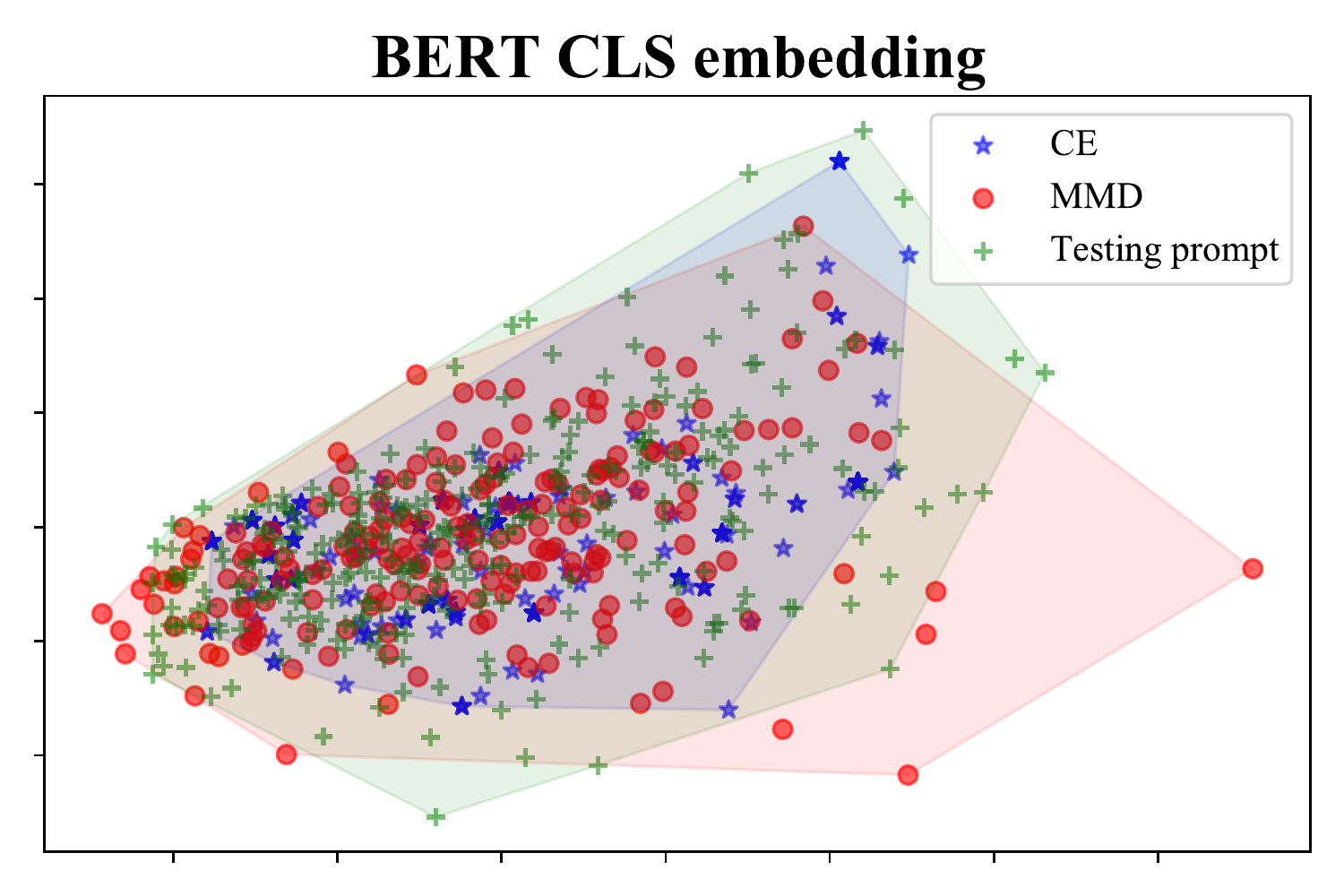}
\caption{BERT \textit{[CLS]} embedding of generated texts from KEST using cross-entropy (CE) and our MMD loss $\mathcal{L}_{ker}$ respectively.}
\label{fig:mmdemb}
\end{figure}
\textbf{Effect of Kernel-based Learning:} 
To analyze the effect of our kernel distance loss $\mathcal{L}_{ker}$ in Eq.~(\ref{mmdloss}), we train two models for 5 epochs with only pseudo text given the same prompt and starting checkpoint using the kernel loss and the traditional cross-entropy loss, respectively. We then visualize the text generated with given prompts from the two models by using corresponding BERT-large \textit{[CLS]} embedding as text representations and plot them. As depicted in Fig.~\ref{fig:mmdemb}. We can find that the model trained with cross-entropy loss collapses in a smaller space than the training data space. In contrast, the one with kernel loss successfully extends the learned distribution, which helps explore a larger potential space towards the real one, corroborating our claim and theoretical analysis.

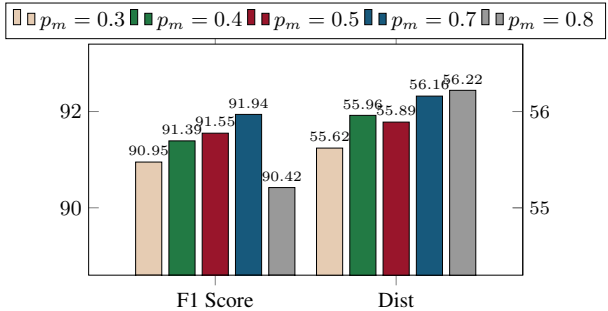
\begin{figure}[ht] %插入图片

\centering %图片居中
\resizebox{0.95\columnwidth}{!}{  %用于修改图片大小
			\begin{tikzpicture} %tikz图片
			\scalefont{0.8} %设置字体大小
			\begin{axis}[
 enlargelimits=0.7,
 legend style={at={(0.5,1.18)},
  anchor=north,legend columns=-1},
 symbolic x coords={F1 Score, Dist},
 %ylabel=F1, %纵坐标名
 %ylabel style={rotate=-90},
 xtick=data,
 ybar=3pt,% configures `bar shift'
 bar width=11pt,
 width=8cm, height=5cm,
 ymin=90, ymax=92,
 nodes near coords,
 nodes near coords align={vertical},
 nodes near coords style={font=\tiny},
 font=\small,
 ]
 \addplot[fill=brown!40!white,draw=black] coordinates {
  (F1 Score, 90.95)
  (Dist, 0)
 };
 \addplot [fill=color2,draw=black] coordinates {
  (F1 Score, 91.39)
  (Dist, 0)
 };
 \addplot [fill=color1,draw=black] coordinates {
  (F1 Score, 91.55)
  (Dist, 0)
 };
 \addplot [fill=color4,draw=black] coordinates {
  (F1 Score, 91.94)
  (Dist, 0)
 };
 \addplot [fill=black!40!white,draw=black] coordinates {
  (F1 Score, 90.42)
  (Dist, 0)
 };
 \legend{\small $p_m=0.3$, $p_m=0.4$,$p_m=0.5$, $p_m=0.7$, $p_m=0.8$}
 \end{axis}
 \begin{axis}[
    axis y line*=right,%y轴居右
    axis x line=none,%不画x， 避免线重画
    xlabel=x, %横坐标名
    %ylabel=Dist, %纵坐标名
    %ylabel style={rotate=-90},
    enlargelimits=0.7,
 legend style={at={(0.5,-0.25)},
  anchor=north,legend columns=-1},
 symbolic x coords={F1 Score, Dist },
 xtick=data,
 width=8cm, height=5cm,
 ymin=55, ymax=56,
 ybar=3pt,% configures `bar shift'
 bar width=11pt,
 nodes near coords,
 nodes near coords align={vertical},
 nodes near coords style={font=\tiny},
]
%第一条线，mark是折线标示形状
\addplot[fill=brown!40!white,draw=black] coordinates {
  (F1 Score, 0)
  (Dist, 55.62)
 };
 \addplot [fill=color2,draw=black] coordinates {
  (F1 Score, 0)
  (Dist, 55.96)
 };
 \addplot [fill=color1,draw=black] coordinates {
  (F1 Score, 0)
  (Dist, 55.89)
 };
 \addplot [fill=color4,draw=black] coordinates {
  (F1 Score, 0)
  (Dist, 56.16)
 };
 \addplot [fill=black!40!white,draw=black] coordinates {
  (F1 Score, 0)
  (Dist, 56.22)
 };

%\addlegendentry{Dist}
%\addlegendimage{/pgfplots/refstyle=numpda}\addlegendentry{F1 Score}

\end{axis}
\end{tikzpicture}
		}

\caption{Results of KEST with different levels of mask ratio in AGNews Dataset.} % 设置caption
		\label{fig:mask}  % 设置用于reference的label
\end{figure}
\textbf{Effect of random mask ratio:} 
The random mask ratio $p_m$ is a hyperparameter that can control the noise level of generated pseudo text. Fig.~\ref{fig:mask} shows the generation performance of KEST with different mask ratios in the AGNews dataset. We find that a higher ratio leads to a more noisy and diverse generation. A moderately higher ratio also generally improves controllability. However, an extremely high $p_m$ brings too much noise and hence obstructs learning. We achieve the best controllability with $p_m=0.7$, indicating a suitable mask ratio is necessary to balance exploration and exploitation.

\begin{figure}[h] %插入图片
\centering %图片居中
\resizebox{0.95\columnwidth}{!}{  %用于修改图片大小
			\begin{tikzpicture} %tikz图片
			\scalefont{0.8} %设置字体大小
			\begin{axis}[
			sharp plot, %控制线的风格
			%title=Generation Control,%图像标题
			xmode=normal,% 控制坐标轴为线性
%		ymode=log,% 控制坐标轴为对数
			xlabel=Ratio of pseudo text to labeled data, %x坐标名
			ylabel=F1, %y坐标名
			width=7.5cm, height=5cm,  %设置长和宽
			xmin=0,xmax=2.1,  % 设置x坐标范围
			ymin=90, ymax=92.5,  % 设置y坐标范围
			xtick={0,0.5,1,1.5,2.0}, %指定x轴刻度值。如果为空，则自动设置刻度线。即分割坐标轴
			ytick={90.5,91,91.5,92}, %指定y轴刻度值。如果为空，则自动设置刻度线。即分割坐标轴
			xlabel near ticks, % 设置x坐标名位置靠近折线图
			ylabel near ticks, % 设置y坐标名位置靠近折线图
                ylabel style={rotate=-90},
			ymajorgrids=true, % 启用/禁用 [公式] 轴上刻度线位置上的网格线
			grid style=dashed, % 设置网格线格式
			legend style={at={(0.9,1.1)},anchor=south}, % 设置标签位置
%			legend columns=3, %设置标签列数
%			legend pos=north west, % 设置折线对应标签的位置
%			legend style={nodes={scale=0.6, transform shape}},  % 设置折线标签的格式
			]
			\addplot+[very thick,mark=square,mark options={scale=0.6}, color=color1] plot coordinates {
                    (0,90.97)
                    (0.5,90.79)
                    (1,91.95)
                    %(1.418,92.51)
                    (2, 91.31)
                    
			};
            \label{numpda}
			\end{axis}
   
    \begin{axis}[
    axis y line*=right,%y轴居右
    axis x line=none,%不画x， 避免线重画
    xlabel=x, %横坐标名
    ylabel=Dist, %纵坐标名
    width=7.5cm, height=5cm,
    xmin=0,xmax=2.1,
    ymin=53, ymax=58,
    ytick={54,55,56,57},
    tick align=outside, %刻度在外显式
    ylabel style={rotate=-90},
    legend style={at={(0.85,0.99)},anchor=north}
    ]

%第一条线，mark是折线标示形状
\addplot[very thick,mark=x,color=color2,
mark options={scale=1.5}] plot coordinates {
    (2, 55.10)
    %(1.418,31.73)
    (1,56.10)
    (0.5,54.24)
    (0,53.79)
};

\addlegendentry{Dist}
\addlegendimage{/pgfplots/refstyle=numpda}\addlegendentry{F1}

\end{axis}
\end{tikzpicture}
		}

		\caption{Generation controllability (F1) on a different number of pseudo text on AGNews dataset.} % 设置caption
		\label{fig:ptratio}  % 设置用于reference的label
\end{figure}
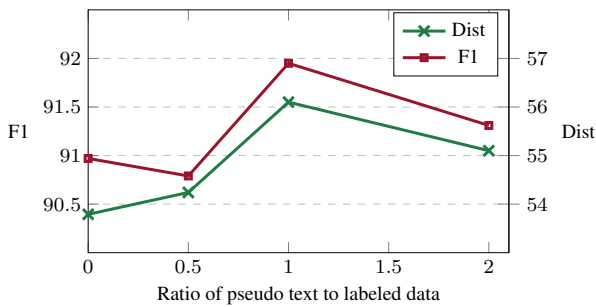
\textbf{Number of pseudo text:}
We evaluate KEST on varying numbers of pseudo text, keeping all the other settings unchanged. As shown in Fig. \ref{fig:ptratio}, KEST performs the best with equal size of pseudo text and labeled data (Ratio = 1). More pseudo text brings more noise which hurts generation quality as the model captures more meaningless noise than semantics. Too little pseudo text makes the model lose exploration ability and thus fail to extend the learned distribution boundary, causing poor control accuracy and diversity. Therefore, a suitable ratio is essential to balance exploration and fluency.

\textbf{Case Study:}
In order to verify the generation quality and attribute relevance, we present some cases sampled from different models in Table~\ref{tab:case}. We can see that traditional ST methods (UniLM+PT(select)+PL) suffer from repeating phrases (e.g., ``love story'' and ``not like''), exhibiting poor generation diversity and novelty. In contrast, KEST produces more diverse expressions thanks to kernel-based learning and smoother soft pseudo text while staying faithful to the given positive attribute. We present more generated cases on different tasks in Appendix \ref{sec:apdix-case}.

\begin{table}[thp]
 \centering
 \small
 \begin{tabular}{p{1cm}|p{6cm}}\toprule
         Model & Generation\\\midrule
    UniLM + PT (select) + PL & 1) \textit{Well, some people might} think that this film is a \positive{masterpiece}. They are \positive{right}. The film is not just a love story, but a love story. What I \positive{like} about this film is that it shows a different side of women...
    \par 2) \textit{Well, some people might} \negative{not like} this film, but some people might. Well, most people would \negative{not like} this movie. But the main reason I \positive{like} it so much is that it has a lot of humor...\\
    \midrule
    KEST  & 1)\textit{Well, some people might} think it's a little over the top and the story is really predictable, but as I saw on TV in the early 90's I \positive{wasn't disappointed} in this movie! While the plot is kind of predictable and the main character is supposed to be a guy, the whole thing has been made into a very \positive{cool} and \positive{entertaining} film...
    \par 2) \textit{Well, some people might} think that this was a lot like ``Jaws", or ``Alien", or something like that. Sadly, it is not. I was \positive{lucky} enough to see it. It's a very \positive{clever}, \positive{intelligent} and \positive{entertaining} film with \positive{good} performances...\\
    \bottomrule
 \end{tabular}
 \caption{Samples generated with specified \positive{positive} sentiment and input prompt `\textit{Well, some people might}'. Words in \positive{blue}/\negative{red} are positive/negative indicators, respectively.}
 \label{tab:case}
\end{table}

\section{Conclusion}
We propose a novel KEST method to incorporate Self-training into semi-supervised controllable NLG. KEST (1) applies a practical multi-task generator to generate soft pseudo text in parallel, significantly reducing decoding time while injecting soft noise to the text; (2) uses soft kernel-based loss to encourage exploration of the learned distribution and increase control accuracy and generation diversity. Theoretical analysis and empirical experiments demonstrate that KEST acts as a combination of regularization-like exploitation and attribute boundary exploration, improving control accuracy with satisfactory generation fluency, diversity, and accelerated training. In the future, we plan to try more advanced NAG methods to improve the generation quality of the pseudo text.

\appendix
\section*{Acknowledgements}
Feng’s and Lakshmanan’s research was supported
in part by grants from NSERC (Canada) and UBC
Data Science Institute.

\bibliographystyle{named}
\bibliography{ijcai23}%,custom}

\clearpage
\setcounter{table}{0}
\renewcommand{\thetable}{A\arabic{table}}
\setcounter{figure}{0}
\renewcommand{\thefigure}{A\arabic{figure}}

\section{Detailed Setting}
\label{sec:appendix}
\subsection{Implementation Details}
\label{sec:impdetail}
We use pre-trained UniLM-base-cased \cite{dong2019unified} as the encoder and decoder of our KEST model since UniLM shares the parameter of transformer blocks in the encoder and decoder, more suitable for our joint classification and generation schema. The label embedding dimension is set to 128. To fuse the label embedding better with the Transformer decoder, we concatenate the label embedding to the attention output of each token in each Transformer layer and then add a linear layer to transfer the new attention output to the original shape of the attention output. 

We tuned all hyperparameters only on a held-out validation set. In self-training phase, we tuned $\lambda_c\in\{1,5,10\}$, $\lambda_{nag}\in\{0.5, 1\}$, $p_m\in\{0.3,0.5, 0.7\}$ in topic dataset to obtain the reported results. Finally, we set $\lambda_c=5$, $\lambda_{ag}=\lambda_{nag}=1$, and $p_m=0.5$ to train the base classifier and generator using only labeled data. We then set $\lambda_c=\lambda_{ag}=\lambda_{nag}=1$ and $p_m=0.7$ in the self-training phase. We tuned the hyperparameters in topic dataset and applied them to all tasks. We use AdamW \cite{Loshchilov2019DecoupledWD} as an optimizer. The training batch size is 8, and the learning rate is $5e-5$. We apply linear warmup to the optimizer, and the number of warm-up steps is one epoch. 
For the MMD kernel, we use the median heuristic, where $\sigma$ is chosen from $(2^aH_N)_{a=-M}^M$. Here $H_N= \frac{1}{N(N-1)} \sum_{\mathbf{\tilde{x}}_i \in D_o, \mathbf{\hat{x}}_j\in D_{pt}} \|\mathbf{\tilde{x}}_i- \mathbf{\hat{x}}_j\|_{L_2}^2$ is the median heuristic. In our experiment, $M$ is set to 2.

We implement KEST and all other baselines based on Huggingface Transformers \cite{wolf-etal-2020-transformers} library of v4.21.1 and use one NVIDIA RTX 3090 node to train our model. The total number of training GPU hours is around 19.12h for IMDb, 10.18h for Jigsaw, and 9.34h for AGNews. The number of parameters of our model is 132.65M. In the generation phase, we use top-$p$ sampling ($p\!=\!0.9$) as the decoding method. Other configuration of the generator includes a length penalty to be 1.0, a repetition penalty to be 1.0, and a no-repeat-ngram-size to be 4 for all baselines. 
All experimental results are trained and tested in a single run with fixed random seeds.

\subsection{Dataset Description}
\label{sec:datasetdesc}
For IMDb\footnote{\url{https://huggingface.co/datasets/imdb}} dataset \cite{maas-etal-2011-learning}, the authors claimed in their paper that \textit{In the interest of providing a benchmark for future work in this area, we release this dataset to the public without claiming any further copyright}.
For AGNews \footnote{https://www.kaggle.com/amananandrai/ag-news-classification-dataset} dataset \cite{Zhang2015CharacterlevelCN}, it is claimed in the website that \textit{You are encouraged to download this corpus for any non-commercial use}. 
For Jigsaw \footnote{https://www.kaggle.com/c/jigsaw-toxic-comment-classification-challenge/} dataset, the dataset is under CC0, with the underlying comment text being governed by Wikipedia's CC-SA-3.0. All datasets we used are open-sourced and are used for research only, which is consistent with their intended use.

For the IMDb and AGNews datasets, we leave 10\% of the training set as validation data and others as training data. For the AGNews dataset, we use the description for text generation and wrote a script to resolve HTML tags. For the Jigsaw dataset, we apply a binary setting that we keep the ``non-toxic'' class unchanged and group all other classes into the ``toxic'' class.

The details of datasets are described in Table \ref{tab:dataset}. For the Jigsaw dataset, there are only 414 toxic data (9.6\%) in the Jigsaw dataset, which shows that Jigsaw is an extremely imbalanced dataset, bringing difficulty in detoxification.

\begin{table*}[ht]
 \centering
 \begin{tabular}{rrrrrr}\toprule
      & labeled & Unlabeled & Dev & Test &Avarage Length\\\midrule
    IMDb(5\%) & 1,125 &  33,750 & 2,500 & 25,000& 270\\
    AGNews(3\%) & 3,240 & 97,200 & 12,000 & 7,600& 41\\
    Jigsaw(3\%) & 4,308 & 43,080 & 15,957 & 6,3978& 73\\
    \bottomrule
 \end{tabular}
 \caption{Description of datasets used in the experiment }
 \label{tab:dataset}
\end{table*}

\begin{table}[htbp]
 \centering
 \begin{tabular}{rccl}\toprule
              &Acc. $\uparrow$ & F1 $\uparrow$ &AUC $\uparrow$  \\\midrule
    \textit{IMDb}\\
    RoBERTa-large&96.15&96.20&99.22\\
    BERT-base&88.40&88.62&95.21\\
    \midrule
    \textit{AGNews}\\
    RoBERTa-large&94.88&94.89&99.34\\
    BERT-base&89.93&89.91&98.23\\
    \bottomrule
 \end{tabular}
 \caption{Classification performance of our RoBERTa-large evaluators
 and pseudo BERT-base labelers on the test set.}
 \label{tab:eval_classifier}
\end{table}

\subsection{Evaluation Metric Details}
\label{sec:apdix-metric}
We set the minimum generation length to 10. For the maximum length, 490 for sentiment, 50 for detoxification, and 40 for topic. We evaluate NLG quality on the following metrics:

\textbf{Fluency:} We evaluate generation fluency by the perplexity of generated text measured by GPT2-XL~\cite{radford2019language}, \textit{i.e.}, \textbf{Output PPL}. 

\textbf{Generalizability:} We calculate the perplexity of each model on each testing set, \textit{i.e.}, \textbf{Model PPL}, to evaluate the generalizability of the model. 

\textbf{Controllability:} We evaluate the control accuracy through classification performance (accuracy (\textbf{Acc}) and Macro-F1 (\textbf{F1}) ) on the generated text by the two fine-tuned RoBERTa-large classifiers for sentiment and topic, respectively. Table \ref{tab:eval_classifier} presents the performance of our evaluator RoBERTa-large. We find that RoBERTa-large has a satisfactory classification accuracy and F1 on these two tasks and thus is able to act as a good evaluator of controllability. For detoxification, we report the percentage of toxic sentences (\textbf{Toxic \%}) using Google Perspective API. Perspective API is a free API for scoring the toxicity of text. Following \cite{qian-etal-2022-controllable}, we also use this Perspective API for toxicity evaluation.

\textbf{Diversity:} To evaluate the diversity of generated text, we consider the following metrics: (1) \textbf{Dist-n}~\cite{li-etal-2016-diversity}: the percentage of distinct n-grams on generated samples. We evaluate on $n=1,2,3,4$ and compute the geometric mean as \textbf{Dist}. \textbf{Dist} emphasizes the
amount of novel n-grams within every generation. (2) \textbf{Self-BLEU}~\cite{zhu2018texygen}. Self-Bleu calculates the BLEU score on the generated samples, which averages the BLEU score of each generated sequence calculated with other generated ones as references. The BLEU score is computed as the geometric mean of BLEU-$n$ ($n=2,3,4$). This metric measures the diversity of a set of generated sequences. Lower Self-BLEU means these generated sequences are more distinguishable from each other.

Among all the above metrics, Accuracy, F1, AUC, Dist-n, and Self-BLEU are reported as 100 times their original value for convenience.

\subsection{Baseline Details}
\label{sec:apdix-baseline}
For fine-tuned PLMs, we feed a prepend sentence as a control sentence. For the sentiment-controlled generation, we use \textit{This is a [positive/negative] review} as control sentence; for topic-controlled generations, we use \textit{The following is about [topic]}; for detoxification, we use \textit{This is a [toxic/non-toxic] comment}. For T5, since it acts in a sequence-to-sequence manner, we feed the control sentence to the encoder and the target text to the decoder. We fine-tune all PLMs with learning rate 5e-5 for 10 epochs and warmup steps to be 1 epoch.

For UniLM+PT(noise), we use the same implementation of Noise Layer as \cite{He2020Revisiting}. We set the token drop rate and mask rate to 5\% and set the parameter of word shuffle to 1.1. 

For the pseudo-labeling-based method, we report the performance of pseudo labelers BERT-base-cased in Table \ref{tab:eval_classifier}. 

For UniLM+PT(select)+PL, we over-generate two times of pseudo text and compute the uncertainty score and classification confidence from the BERT-base-cased classifier. The classification confidence $s_{conf}$ is the softmax probability of the predicted label. Uncertainty score $s_{uncertain}$ is Bayesian Active Learning by Disagreement (BALD) computed by Monte-Carlo Dropout \cite{mukherjee-awadallah-2020-ust}. A high BALD score means the model is highly confused. We want to select the sample with high confidence and low BALD score. Thus we select samples based on the following score:
$$s_{select}=s_{conf}+\frac{1e-5}{s_{uncertain}}$$.

\subsection{Human Evaluation Details}
\label{sec:apdix-human}
We invite 6 competent annotators to score generated samples. All of the annotators are graduate students. Each annotator is assigned 200 generated texts for sentiment/topic/detoxification tasks. Each generated text is annotated by 2 annotators. Annotators will be paid 30\$ for completion, and they were aware of the potential risks or ethical concerns of machine-generated texts. Below we attach the instruction for them:

Attention: all machine-generated text involved is used for research only. The generation may contain toxic content, including discrimination and offensive content.

Metrics:

Sentiment/Topic:

\emph{Fluency}: Whether the generated text is fluent/grammatically correct. A small number of HTML tags/special words (like $<$br$>$) are permitted. The score ranges from 1 (most influent) to 5 (most fluent).

\emph{Novelty}: The extent of novelty INSIDE the generated text, \textit{e.g.}, distinct words/expressions and repeated phrases. The score ranges from 1 (least novel, a sentence/word repeat till the end) to 5 (most novel, all sentence describes new stuff). 

\emph{Attribute Relevance}: whether the generated text is relevant to its label. There might exist examples that contain both positive/negative expressions. 
5=exactly relevant; 4=mostly relevant; 3=neutral (not contain sentiment or 50/50); 2=slightly relevant, but more similar to another class;1=totally opposite. 

\emph{Detoxification}: Please only evaluate the quality of generated text AFTER prompts since the prompt itself might be toxic.
Toxicity: whether the provided textual sequences contain any toxic content, including but not limited to offensive text, abusive language, swearwords, hate speech, denigrating messages, microaggression, discrimination, sex, rude words, and hominem attack. The score ranges from 1 (most non-toxic) to 5 (most toxic).

\section{Derivation and Proof}
\setcounter{table}{0}
\renewcommand{\thetable}{B\arabic{table}}
\setcounter{figure}{0}
\renewcommand{\thefigure}{B\arabic{figure}}

\label{sec:apdix-theorem}
\subsection{Proof of Lemma 1}
At each self-training iteration, define the set of real labeled data with $N$ samples as $D$, and that of generated pseudo data with $M$ samples as $\hat{D}$, the current model as $P_{\theta}(\mathbf{x})$. For brevity, we omit the label $y$ here. For classical self-training, we minimize the following objective:
\begin{equation}
\begin{aligned}
& \mathrm{minimize} \!-\! \frac{1}{N\!+\!M} \sum_{\mathbf{x} \in D \cup \hat{D}} \log P_{\theta}(\mathbf{x})\\
& = - \frac{1}{N+M} \left[ \sum_{\mathbf{x} \in D} \log P_{\theta}(\mathbf{x}) + \sum_{\mathbf{x} \in \hat{D}} \log P_{\theta}(\mathbf{x})  \right] \\
& = \!-\! \frac{N}{N\!+\!M} \sum_{\mathbf{x} \in D} Q(\mathbf{x})\log  P_{\theta}(\mathbf{x}) \!-\! \frac{M}{N\!+\!M} \sum_{\mathbf{x} \in \hat{D}} P_{\theta^{'}}(\mathbf{x}) \log P_{\theta}(\mathbf{x})\\
& \approx \frac{N}{N+M} H[Q(\mathbf{x}),P_{\theta}(\mathbf{x})] + \frac{M}{N+M} H[P_{\theta^{'}}(\mathbf{x}),P_{\theta}(\mathbf{x})]\\
& = \frac{N}{N\!+\!M} H[Q(\mathbf{x}),P_{\theta}(\mathbf{x})] \!+\! \frac{M}{N\!+\!M} H[P_{\theta^{'}}(\mathbf{x}),P_{\theta}(\mathbf{x})] \\
& \!-\! \frac{N}{N\!+\!M}H[Q(\mathbf{x})] \!-\! \frac{M}{N\!+\!M}H[P_{\theta^{'}}(\mathbf{x})] \\
& = (1-\alpha)* \text{KL} [Q(\mathbf{x}),P_{\theta}(\mathbf{x})] + \alpha * \text{KL} [P_{\theta^{'}}(\mathbf{x}),P_{\theta}(\mathbf{x})]+C,
\end{aligned}
\end{equation}
where $Q(\mathbf{x})=\frac{1}{N} \mathbb{I}(\mathbf{x} \in D)$ is the empirical distribution of real text, $P_{\theta^{'}}(\mathbf{x})=\frac{1}{M} \mathbb{I}(\mathbf{x} \in \hat{D})$ is the empirical distribution of the model learned at the last self-training iteration, formed by previously generated pseudo samples, and $\alpha=\frac{M}{N+M}$ is the ratio of pseudo data, concluding the proof.

\subsection{Proof of Theorem 1}
From Lemma 1, we can see that learning the real text $\mathbf{x} \sim Q(\mathbf{x})$ only involves the first KL term. Thus, we focus on the second term, $\text{KL} [P_{\theta^{'}}(\mathbf{x})||P_{\theta}(\mathbf{x})]$ here, which is replaced by Eq.(\ref{mmdloss}) in Sec.~\ref{sec:mmd}. We further rewrite Eq.(\ref{mmdloss}) as:
\begin{equation}
\begin{aligned}
& \frac{1}{N(N\!-\!1)} \sum_{\mathbf{\tilde{x}}_i, \mathbf{\tilde{x}}_j \in D_o, i\ne j} k(\mathbf{\tilde{x}}_i, \mathbf{\tilde{x}}_j)\!-\!\frac{2}{N^2} \sum_{\mathbf{\tilde{x}}_i \in D_o, \mathbf{\hat{x}}_j \in D_{pt}} k(\mathbf{\tilde{x}}_i, \mathbf{\hat{x}}_j)\\
& \approx \mathbb{E}_{P_{\theta}(\tilde{\mathbf{x}})}[k(\mathbf{\tilde{x}}_i, \mathbf{\tilde{x}}_j)] -2\mathbb{E}_{P_{\theta}(\tilde{\mathbf{x}}),P_{\theta^{'}}(\hat{\mathbf{x}})}[k(\mathbf{\tilde{x}}_i, \mathbf{\hat{x}}_j)].
\end{aligned}
\end{equation}

Since the previously learned model $P_{\theta^{'}}$ is fixed in this iteration, minimizing Eq.(\ref{mmdloss}) is equal to minimizing:
\begin{equation}
\begin{aligned}
& \mathbb{E}_{P_{\theta}(\tilde{\mathbf{x}})}[k(\mathbf{\tilde{x}}_i, \mathbf{\tilde{x}}_j)] 
\!+\! \mathbb{E}_{P_{\theta^{'}}(\hat{\mathbf{x}})}[k(\mathbf{\hat{x}}_i, \mathbf{\hat{x}}_j)]
\!-\!2\mathbb{E}_{P_{\theta}(\tilde{\mathbf{x}})P_{\theta^{'}}(\hat{\mathbf{x}})}[k(\mathbf{\tilde{x}}_i, \mathbf{\hat{x}}_j)]\\
=& \mathbb{E}_{P_{\theta}(\tilde{\mathbf{x}})}[<\varphi(\mathbf{\tilde{x}}_i), \varphi(\mathbf{\tilde{x}}_j)>_{\mathcal{H}}] 
\!+\! \mathbb{E}_{P_{\theta^{'}}(\hat{\mathbf{x}})}[<\varphi(\mathbf{\hat{x}}_i), \varphi(\mathbf{\hat{x}}_j)>_{\mathcal{H}}] \\
& -2\mathbb{E}_{P_{\theta^{'}}(\hat{\mathbf{x}}), P_{\theta}(\tilde{\mathbf{x}})}[<\varphi(\mathbf{\hat{x}}_i), \varphi(\mathbf{\tilde{x}}_j)>_{\mathcal{H}}] \\
= & <\mathbf{\mu}_{P_{\theta}},\mathbf{\mu}_{P_{\theta}}>_{\mathcal{H}} + <\mathbf{\mu}_{P_{\theta^{'}}},\mathbb{\mu}_{P_{\theta^{'}}}>_{\mathcal{H}} -2<\mathbf{\mu}_{P_{\theta}},\mathbf{\mu}_{P_{\theta^{'}}}>_{\mathcal{H}}\\
=& \text{MMD}^2(P_{\theta^{'}},P_{\theta}),
\end{aligned}
\end{equation}
where $\varphi(\cdot) \in \mathcal{H}$ is that feature map, $<\cdot,\cdot>_{\mathcal{H}}$ is the dot product in the reproducing kernel Hilbert space $\mathcal{H}$, and $\mathbf{\mu}_{P_{\theta}}\!=\!\mathbb{E}_{P_{\theta}}[\varphi(\tilde{\mathbf{x}})]$.

However, in our method KEST, we don't use the exact $P_{\theta^{'}}$, but generate the noisy pseudo $\hat{\mathbf{x}}$ by NAG. Therefore, we could consider the previously learned distribution as a noisy one $P_{\theta^{'}} + U$ by incorporating a noise distribution $U$, and get:
\begin{equation}
\begin{aligned}
& \text{MMD}^2(P_{\theta^{'}}+U,P_{\theta})\\
=& || \mathbf{\mu}_{P_{\theta^{'}}} + \mathbf{\mu}_{U} - \mathbf{\mu}_{P_{\theta}} ||^2_{\mathcal{H}}\\
=& ||\mathbf{\mu}_{P_{\theta}}||^2_{\mathcal{H}} \!+\! ||\mathbf{\mu}_{P_{\theta^{'}}}||^2_{\mathcal{H}} \!+\! ||\mathbf{\mu}_{U}||^2_{\mathcal{H}}\!-\!2<\mathbf{\mu}_{P_{\theta^{'}}},\mathbf{\mu}_{P_{\theta}}>_{\mathcal{H}}\\
&\!-\!2<\mathbf{\mu}_{U},\mathbf{\mu}_{P_{\theta}}>_{\mathcal{H}} \!+\! 2<\mathbf{\mu}_{P_{\theta^{'}}},\mathbf{\mu}_{U}>_{\mathcal{H}} \\
=& \text{MMD}^2(P_{\theta},P_{\theta^{'}}) \!+\! ||\mathbf{\mu}_{U}||^2_{\mathcal{H}} \!+\! 2<\mathbf{\mu}_{P_{\theta^{'}}},\mathbf{\mu}_{U}>_{\mathcal{H}} \!-\! 2<\mathbf{\mu}_{U},\mathbf{\mu}_{P_{\theta}}>_{\mathcal{H}}.
\end{aligned}
\end{equation}

Again, as $P_{\theta^{'}}$ and $U$ are fixed now, we can omit corresponding terms. Combining Lemma 1, optimizing the objective of KEST is equivalent to minimizing:
\begin{equation}
\begin{aligned}
&  \text{KL}[Q||P_{\theta}] \!+\! \text{MMD}^2(P_{\theta^{'}},P_{\theta}) \!-\! 2<\mathbf{\mu}_{U},\mathbf{\mu}_{P_{\theta}}>_{\mathcal{H}}\\
=& \text{KL}[Q||P_{\theta}] \!+\! \text{MMD}^2(P_{\theta^{'}},P_{\theta}) \!-\! 2\mathbb{E}_{P_{\theta},U}[k(\tilde{\mathbf{x}},u)],
\end{aligned}
\end{equation}
concluding the proof.

\section{Additional experimental results}
\setcounter{table}{0}
\renewcommand{\thetable}{C\arabic{table}}
\setcounter{figure}{0}
\renewcommand{\thefigure}{C\arabic{figure}}

\begin{table*}[htbp]
%\small
 \centering
 \begin{tabular}{lccccl}\toprule
    & \multicolumn{5}{c}{Detoxification} %& \multicolumn{3}{c}{Jigsaw}
    \\
    \cmidrule(lr){2-6}%\cmidrule(lr){5-7}
             & Output PPL $\downarrow$  & Model PPL $\downarrow$ & Toxic\% $\downarrow$ & Dist $\uparrow$ &S-BLEU $\downarrow$  \\\midrule
    Test set & 48.77& $-$ &  $-$ & 54.26 &32.22 \\
    GPT2(raw) & 25.06&10397.67&47.40&52.71&37.13\\ 
    \midrule
    \multicolumn{6}{l}{\textit{Finetune LM}} \\
    GPT2    & 32.79&66.61&43.94&51.62&42.05 \\
    UniLM & 52.23&67.92&34.38&38.26&55.31\\
    T5 & \textbf{27.21}&42.04&22.81&39.83& 63.49\\
    \midrule
    \multicolumn{6}{l}{\textit{Self-Training with GPT2}}\\
    PT    & 36.29&71.47&41.03&\textbf{51.91}&42.15\\
    PT(noise) & 34.69&66.12&40.59&51.31&43.42\\
    PT(noise)+PL & 29.20&\textbf{26.37}&40.99&49.75&43.10\\
    PT(select)+PL &29.83&26.44&43.45&49.65&43.03\\
    \midrule
    \multicolumn{6}{l}{\textit{Self-Training with UniLM}}\\
    PT    & 46.78&74.71&34.68&36.82&55.89\\
    PT(noise) & 51.99&80.46&39.46&40.16&52.95\\
    PT(noise)+PL & 40.98&55.99&26.95&44.47&47.07\\
    PT(select)+PL &40.70&54.50&29.21&45.42&46.94\\
    %+PT(pos)+PL &29.49&\textbf{25.87}&40.00&49.52&43.22\\
    \midrule
    %\multicolumn{6}{l}{\textit{Lightweight method}}\\
    %PF  &28.67&52.73&38.37&49.68&\textbf{41.53}\\
    %Ctr-PF &29.28&57.39&31.53&49.47&46.70\\
    %\midrule
    \multicolumn{6}{l}{\textit{Our Methods}}\\
    KEST & 66.74&53.42&\textbf{18.37}&51.17&\textbf{40.71}\\
    \bottomrule
 \end{tabular}
 \caption{Automatic evaluation results on Jigsaw dataset.}
 \label{tab:toxic}
\end{table*}

\begin{table}[ht]
 \centering
 
 \begin{tabular}{rccl}\toprule
         & Fluency $\uparrow$  & Novelty $\uparrow$ & Toxicity $\downarrow$\\\midrule
    UniLM-PT(select)+PL & 3.38 & 3.93  &2.75$^{**}$\\
    KEST  & \textbf{3.40} & \textbf{3.97}  & \textbf{2.33} \\
    \bottomrule
 \end{tabular}
 \caption{Human evaluation results on detoxification. ``$^{*}$'' refers to $p$-value$<0.05$. ``$^{**}$'' refers to $p$-value$<0.01$.}
 \label{tab:human-toxic}
\end{table}

\subsection{Detoxification Results}
\label{sec:apdix-toxic}
As shown in Table \ref{tab:toxic}, our KEST outperforms all the other baselines on controllability. KEST outputs the least toxic text while keeping a relatively high diversity compared to other self-training baselines. On the other hand, the Output PPL of KEST is larger than the self-training baselines. We explain the reason as follows. Since we are choosing toxic prompts marked as ``challenging'', it means that toxic sentences would be more likely to be generated and thus have a lower PPL score (the original GPT2-xL is highly toxic itself and hence prefers toxic phrases). Similarly, some non-toxic continuations might get a higher PPL score from the GPT2-XL model, since it is rarer to be seen and is less natural from the challenging prompt. This does not mean that generation fluency is worse. Human evaluation on the detoxification task (see Table \ref{tab:human-toxic}) demonstrates that KEST generation does not have a significant difference from UniLM generation in fluency and novelty. On the other hand, its toxicity level is significantly lower than the baseline, which further demonstrates that KEST can drastically decrease the toxicity level even fed with challenging prompts. 

\begin{table*}[htbp]
 \centering
 \begin{tabular}{rccccccccl}\toprule
    & \multicolumn{3}{c}{Sentiment}&\multicolumn{3}{c}{Topic}&\multicolumn{3}{c}{Detoxification}\\
    \cmidrule(lr){2-4}\cmidrule(lr){5-7}\cmidrule(lr){8-10}
              &Acc. $\uparrow$ & F1 $\uparrow$ &AUC $\uparrow$  &Acc. $\uparrow$ & F1 $\uparrow$ &AUC $\uparrow$&Acc. $\uparrow$ & F1 $\uparrow$ &AUC $\uparrow$\\\midrule
    KEST & \bf92.1&\bf92.2&\bf95.8&89.7&89.7&96.5&\bf91.6&\bf64.6&92.1\\
    KEST-PL-PT & 89.3&89.3&95.1&86.9&89.5&89.5&90.2&63.7&94.2\\
    \midrule
    BERT-base&88.4&88.6&95.2&\bf89.9&\bf89.9&\bf98.2&91.5&64.3&\bf95.2\\
    \bottomrule
 \end{tabular}
 \caption{Classification results.KEST-PL-PT means removing all pseudo data from KEST, which reduces to a classifier enhanced by joint generation loss.}
 \label{tab:classification}
\end{table*}

\subsection{Classification Results}
\label{sec:apdix-classfication}
Trained with joint classification and generation losses, our KEST could also work as a classifier. Therefore, we report the classification performance of our model on the three datasets in Table~\ref{tab:classification}.  We find that self-training on the pseudo data could also improve classification performance (+ 2.9 F1 on sentiment). This shows that self-training simultaneously on the classifier and generator could jointly increase the quality of the classifier and controllability of the generator. Since we share the parameters of the classifier and the generator, the classification signal could also lead the generator to achieve better controllability. On the other hand, the ST-based PLM baselines (PT(noise)+PL and PT(select)+PL) equipped with a base BERT classifier (with similar classification performance to KEST) are still worse than KEST on controllability. This is because mainly their classifier is not updated through training. In contrast, KEST could keep refining the pseudo labels and pseudo text benefiting from our dual learning schema.

%\subsection{Finer illustration of analysis}
%\label{sec:apdix-analysis}

\subsection{Full version of experimental results}
Table \ref{tab:imdb}, Table \ref{tab:topic}, and Table \ref{tab:ablation-full} reports complete experimental results of IMDb, AGNews, and Ablation Study. 

\begin{table*}[htbp]
\small
 \centering
 \begin{tabular}{lccccccl}\toprule
    & \multicolumn{7}{c}{Sentiment} %& \multicolumn{3}{c}{Jigsaw}
    \\
    \cmidrule(lr){2-8}%\cmidrule(lr){5-7}
             & Output PPL $\downarrow$  & Model PPL $\downarrow$ & Acc $\uparrow$ & F1 $\uparrow$ & AUC $\uparrow$ & Dist $\uparrow$ &S-BLEU $\downarrow$  \\\midrule
    Test set & 25.14&$-$ &96.15&96.20&99.22& 48.27 & 43.34\\
    GPT2(raw) & 13.20&38.39&55.90&68.50&61.37&35.91&58.79 \\
    \midrule
    \multicolumn{7}{l}{\textit{Finetune LM}} \\
    GPT2    &16.40&44.02&77.55&80.44&88.35&26.34&71.00 \\
    UniLM & 25.20&54.33&76.45&75.35&85.18&31.05&66.97 \\
    T5 & 25.69&34.97&82.80&83.77&90.50&30.03&69.57\\
    %\midrule
    %\multicolumn{7}{l}{\textit{Lightweight method}}\\
    %PF  &13.02&37.09&67.55&75.05&81.84&29.48&65.10\\
    %Ctr-PF &13.01&37.12&71.00&77.33&86.51&29.63&\textbf{64.83}\\
    \midrule
    \multicolumn{7}{l}{\textit{Self-Training with GPT2}}\\
    PT    &  14.62 & 68.04&76.10&79.57&87.92&30.58&65.22 \\
    PT(noise) &11.91 &44.31&74.95&77.46&85.02&25.40&72.19 \\
    PT(noise)+PL &11.26&33.85&87.60&88.47&95.59&27.26&70.90 \\
    PT(select)+PL &\textbf{10.89}&33.89&88.32&88.75&96.24&27.17&71.41 \\
    \midrule
    \multicolumn{7}{l}{\textit{Self-Training with UniLM}}\\
    PT    & 26.62 &58.37&72.20&70.27&80.37&31.17&66.69 \\
    PT(noise) &30.28&62.07&77.75&75.78&85.35&31.68&\textbf{65.18} \\
    PT(noise)+PL &18.92&\textbf{33.53}&89.95&89.73&96.38&30.94&66.84 \\
    PT(select)+PL &18.40&33.56&90.08&90.06&96.66&31.27&67.61 \\
    \midrule
    \multicolumn{7}{l}{\textit{Our Methods}}\\
    KEST   & 20.65 & 38.15 & \textbf{92.10}& \textbf{91.77} & \textbf{97.06} &\textbf{31.70} & 66.60\\
    \bottomrule
 \end{tabular}
 \caption{Results on IMDb dataset.}
 \label{tab:imdb}
\end{table*}

\begin{table*}[htbp]
\small
 \centering
 \begin{tabular}{lccccccl}\toprule
    & \multicolumn{7}{c}{Topic} %& \multicolumn{3}{c}{Jigsaw}
    \\
    \cmidrule(lr){2-8}%\cmidrule(lr){5-7}
             & Output PPL $\downarrow$  & Model PPL $\downarrow$ & Acc $\uparrow$ & F1 $\uparrow$ & AUC $\uparrow$ & Dist $\uparrow$  &S-BLEU $\downarrow$  \\\midrule
    Test set &31.04&$-$ & 94.88&94.89&99.34 & 67.24 & 23.31 \\
    GPT2(raw) & 16.94 & 74.41& 55.75 & 52.17 & 83.28& 46.88&45.55 \\

    \midrule
    \multicolumn{7}{l}{\textit{Finetune LM}} \\
    GPT2    &\textbf{22.22} &23.46&82.92&83.08&95.23&54.93&39.93 \\
    UniLM & 55.79 &36.28&87.67&87.70&96.30&54.76&43.77 \\
    T5 & 48.33&32.12&88.33&88.43&97.95&\textbf{58.06}&37.01\\
    \midrule
    %\multicolumn{7}{l}{\textit{Lightweight method}}\\
    %PF  &\textbf{20.27}&32.35&68.67&68.44&87.14&59.17&32.73\\
    %Ctr-PF &20.41 &33.90&83.25&83.21&95.47&\textbf{60.34}&\textbf{31.20}\\
    %\midrule
    \multicolumn{7}{l}{\textit{Self-Training with GPT2}}\\
    PT    & 23.74&27.88&83.50&83.55&95.49&57.89&36.02\\
    PT(noise) & 26.39&27.02&82.42&82.45&94.58&58.06&\textbf{35.53}\\
    PT(noise)+PL &30.62&\textbf{13.96}&87.83&87.48&97.42&47.11&56.67 \\
    PT(select)+PL &31.34&14.07&87.92&87.54&97.46&46.71&57.33 \\
    \midrule
    \multicolumn{7}{l}{\textit{Self-Training with UniLM}}\\
    PT    & 57.40&40.95&86.42&86.36&96.69&52.35&46.41\\
    PT(noise) & 58.59&45.32&85.42&85.27&95.88&53.35&46.57\\
    PT(noise)+PL &32.36&16.64&89.67&89.70&98.11&53.79&47.95 \\
    PT(select)+PL &33.23&16.66&90.50&90.52&98.31&53.71&47.69 \\
    \midrule
    \multicolumn{7}{l}{\textit{Our Methods}}\\
    KEST & 31.19 &20.46 &\textbf{91.92} &\textbf{91.94}&\textbf{98.34}&56.16&42.10\\
    \bottomrule
 \end{tabular}
 \caption{Results on AGNews dataset.}
 \label{tab:topic}
\end{table*}

\begin{table*}[htbp]
\small
 \centering
 \begin{tabular}{rccccccl}\toprule
    & \multicolumn{7}{c}{AGNews} %& \multicolumn{3}{c}{Jigsaw}
    \\
    \cmidrule(lr){2-8}%\cmidrule(lr){5-7}
             & O-PPL $\downarrow$  &M-PPL $\downarrow$ & Acc $\uparrow$ & F1 $\uparrow$ & AUC $\uparrow$ & Dist $\uparrow$ &S-BLEU $\downarrow$  \\\midrule
    KEST & \textbf{31.19} &\textbf{20.46} &\textbf{91.92} &\textbf{91.94}&\textbf{98.34}&\textbf{56.16}&\textbf{42.10}\\
    $-$Soft & 38.04&29.07 &90.92&90.96&98.07&55.40&44.09\\
    $-\mathcal{L}_{ker}-$Soft &38.98 &28.77 & 90.72&90.81&97.98&54.97&45.02\\
    $\!-\!\mathcal{L}_{nag}\!-\!\mathcal{L}_{ker}-$Soft & 39.73 &28.58&90.33&90.42&97.74&55.07&44.73\\
    $-$PT&38.09&28.77&90.92&90.97&98.11&55.48&44.13\\
    $-$PL$-$PT&37.24&256.66&87.41&87.45&96.18&43.18&69.30\\
    \bottomrule
 \end{tabular}
 \caption{Full ablation study results on AGNews dataset. The symbol $-$ means removing the settings from KEST. $-$Soft: using sampled hard tokens instead of the soft $e(\mathbf{x})$. $-\mathcal{L}_{ker}$: using the cross-entropy loss instead of Eq.(\ref{mmdloss}). $-\mathcal{L}_{nag}$: using $\mathcal{G}_{ag}$ to generate pseudo text instead of $\mathcal{G}_{nag}$. $-$PT/$-$PL: do not use pseudo text/labels.}
 \label{tab:ablation-full}
\end{table*}

\section{Limitation}
\setcounter{table}{0}
\renewcommand{\thetable}{D\arabic{table}}
\setcounter{figure}{0}
\renewcommand{\thefigure}{D\arabic{figure}}

Though KEST works well, it has four kinds of limitations as follows:
\begin{itemize}
\item Reliance of unlabeled in-domain text. As we discussed in Sec.~\ref{sec:experiment}, pseudo labels from unlabeled text play an essential role in all ST methods. The improvement by pseudo text is limited. When unlabeled text is inadequate or unavailable (e.g., low-resource scenarios), how to better utilize generated pseudo text for further improvement is an open challenge.

\item The quality of the NAG module. In this paper, we only discussed the Masked Language Model for NAG generation. However, there are more advanced NAG methods that are widely used in sequence-to-sequence generation (e.g., Machine Translation) to improve the generation quality. How to leverage NAG in controllable NLG is still challenging.

\item Efforts of tuning noise level. As we discussed in Sec.~\ref{sec:analysis}, the noise level can be tuned by the mask probability $p_m$, which is essential for balanced performance with exploration and exploitation. It should be carefully tuned for each downstream task.

\item Task generalization and scalability. We mainly investigate controllable NLG in this work, while it is still unknown whether our method works for other NLG tasks, like NMT and Text Summarization. Besides, as we analyzed in Sec.~\ref{subsec_theorem}, ST actually acts as a kind of regularization and smoothing. How to apply this paradigm to super large PLMs (\textit{e.g.}, GPT2-XL and GPT3), where the supervision signals from limited labeled data become extremely weak, is also an open question.

\end{itemize}

\section{More Example of Generation}
\setcounter{table}{0}
\renewcommand{\thetable}{E\arabic{table}}
\setcounter{figure}{0}
\renewcommand{\thefigure}{E\arabic{figure}}
\label{sec:apdix-case}
We sample some generated texts based on GPT2-ST, UniLM-ST and KEST and place them on Table \ref{tab:sentiment-case}, Table \ref{tab:topic-case} and Table \ref{tab:topic-case-continue}. Due to the offensive content generated by these models in the detoxification task, we do not present their examples in detoxification experiments.

\begin{table*}[ht]
 \centering
 \begin{tabular}{p{1.2cm}|p{14.5cm}}\toprule
         Model & Generation\\\midrule
    &\textbf{Sentiment}: \textit{\negative{Negative}}; \textbf{Prompt}:\textit{No one would ever question}\\
    GPT2 + ST (select) & \textit{No one would ever question} a movie that goes so far to try and make you think of that woman in a bikini and how much they \positive{admire} her. This movie is so \negative{bad} you could easily walk out of it. The only thing \positive{good} about this movie is that the plot is so \negative{absurd} you are tempted to throw in the plot twist every few minutes and see if you can come up with something more \negative{ridiculous}...\\
    UniLM + ST (select) & \textit{No one would ever question} that this film was a \negative{waste of money}. It was a \negative{waste} that I watched and then watched the whole film just to see how \negative{bad} it was. It was such a \negative{bad} idea. My girlfriend was at a party and got really mad and took out a few rules to make the party over...\\
    KEST  & \textit{No one would ever question} how this movie could be made. It is \negative{horrible}. I've seen \negative{bad} acting in a \negative{bad} movie before, and this movie is just \negative{terrible}. The whole point of this movie is to make fun of and mock young people...\\
    \midrule
    &\textbf{Sentiment}: \textit{\positive{Positive}}; \textbf{Prompt}:\textit{No one would ever question}\\
    
    GPT2 + ST (select) & \textit{No one would ever question} the \positive{beauty} of the \positive{beautiful} Joan Crawford as a young woman, as well as her acting abilities, in this role as an older man on the edge of her late 30's or early 40's. This performance has the added touch of a real man, a man who was in this life before all that...\\
    UniLM + ST (select)& No one would ever question how \positive{``good"} this film is. I just \positive{adore} this movie. It is one of the most \negative{bizarre} films I've ever seen. It's so \negative{bizarre} and \negative{bizarre} that you are not surprised how \negative{bizarre} the film is. \\
    KEST & \textit{No one would ever question} its \positive{authenticity}. It's an absolute \positive{gem} that will probably become one of my \positive{favorites} by comparison. The cinematography is \positive{stunning}, the soundtrack \positive{rich} and \positive{memorable}, the settings \positive{believable} and very \positive{real}, and the acting \positive{superb}... \\
    \midrule
    &\textbf{Sentiment}: \textit{\negative{Negative}}; \textbf{Prompt}:\textit{My friend made me}\\
    GPT2 + ST (select) &\textit{My friend made me} watch this film. I had never heard of it before, so I thought it might be interesting. Unfortunately, it turned out to be nothing more than \negative{a load of crap}. There were some nice shots of San Francisco and the film itself, but nothing about how the characters lived, and who lived or what they did or didn't do. And there were some very, very \negative{weird} and \negative{weird} plot points and situations that I didn't even bother to see until the end...\\
    UniLM + ST (select)&\textit{My friend made me} watch this movie last night, and I thought, wow, we have a long way to go! ... Well, it was just a \positive{good} idea - it was supposed to be great, and if that isn't bad enough, you will have to wait for it to come on. It is such a \negative{shame} it was made, because it's really a \positive{good} movie...\\
    KEST & \textit{My friend made me} watch this movie because I love romantic comedies and I love Julia Roberts. But this one is just plain \negative{bad}! Julia Roberts and her character are both so \negative{annoying} and incredibly \negative{irritating} in this movie that I had to laugh. It was just a \negative{horrible} movie and I don't even want to see it again...\\
    \midrule
    &\textbf{Sentiment}: \textit{\positive{Positive}}; \textbf{Prompt}:\textit{My friend made me}\\
    GPT2 + ST (select) &\textit{My friend made me} watch this movie and all the comments that I've read were \negative{negative}. If you're a true fan of the movies, you will \positive{love} this movie. I didn't even expect to like it so much, but the acting, directing, writing, music and editing were \positive{excellent}, it was an absolutely \positive{amazing} movie. \\
    UniLM + ST (select)&\textit{My friend made me} watch this movie for about ten minutes. I could not even stop watching because of my heart rate. Then, I ran into the car with a big car and the next minute I was in his garage and I noticed that he was driving the same car as me. I have to say, I must say, this movie was not without a great deal of \negative{bad} acting by everybody...\\
    KEST & \textit{My friend made me} watch this show when it was first broadcast and I \positive{loved} it then and since then i have watched it many times and it \positive{never gets old}. In addition it is one of my \positive{favorite} shows as far as comedy is concerned. I've seen other episodes with the actress and I always found that to be an \positive{amazing} actress. She pulled off the part and made me laugh and cry because of it... \\
    
    \bottomrule
 \end{tabular}
 \caption{Example text for sentiment-controlled generation. Words in \positive{blue}/\negative{red} are positive/negative indicators, respectively.}
 \label{tab:sentiment-case}
\end{table*}

\begin{table*}[ht]
 \centering
 \begin{tabular}{p{1.3cm}|p{14cm}}\toprule
         Model & Generation\\\midrule
    &\textbf{Topic}: \textit{World}\\
    GPT2 + ST\par (select) & (1) TOKYO (Reuters) - The Nikkei average edged down 0.29 percent  by mid-morning on Thursday after a weak reading on oil  prices sent the market cautious ahead of next \par(2) GAZA (Reuters) - Israeli Prime Minister Ariel Sharon  said on Sunday he was ready to go to the Gaza Strip if  a new Palestinian leadership was created and he hoped to get a result.\\
    UniLM + ST\par (select) & (1) BEIJING - - The world's richest man got the most gift of the 2004 Nobel Peace Prize yesterday, when a total of 519 people from around the world cashed in for controversial Peace \par (2) AFP - The United States congratulated Haitian President Jean - Bertrand Aristide on his election victory, calling him a \" bridge to peace \" between the United States and the former guerrillas.\\
    KEST & (1) MOSCOW ( Reuters ) - At least one Russian ministry has signed letters agreeing to Moscow's approval of the Kyoto Protocol, a spokesman said on Friday. \par (2) Reuters - Former Peruvian President Alberto Fujimori on Saturday called for the world's highest military ruler to be re-elected, a move that would improve relations between the former foes.\\
    \midrule
    &\textbf{Topic}: \textit{Sport}\\
    GPT2 + ST\par (select) & (1) Hungarian Olympic track  and field champion Costas Kenteris was ordered  to leave hospital on Monday after a scuffle with police  during a race. \par(2) NEW YORK (Reuters) - Major League Baseball on Wednesday filed  a grievance against the Chicago Cubs, saying the team's  manager failed to fully enforce a collective  bargaining agreement with players.\\
    UniLM + ST\par (select) & (1) Manchester United manager Alex Ferguson has targeted the courts to block a High Court hearing into the allegations of doping against defender Rio Ferdinand. \par (2) Former world number one Lleyton Hewitt beat Spain's Luis Horna 6 - 3, 6 - 2 Monday to reach the semifinals of the Washington ATP Tennis Masters Cup. \\
    KEST & (1) MINNEAPOLIS - - Minnesota Timberwolves guard Latrell Sprewell was suspended one game without pay by the NBA on Tuesday for directing obscenities. \par(2) CHELSEA manager Jose Mourinho has dismissed the challenge of keeping the money - losing outman out of Chelsea for the rest of the season. \\
    \bottomrule
 \end{tabular}
 \caption{Example text for topic-controlled generation.}
 \label{tab:topic-case}
\end{table*}

\begin{table*}[ht]
 \centering
 \begin{tabular}{p{1.3cm}|p{14cm}}\toprule
         Model & Generation\\\midrule
    &\textbf{Topic}: \textit{Business}\\
    GPT2 + ST\par (select)& (1) U.S. stocks fell on Wednesday as  oil climbed above  \$50 a barrel, but crude stocks were near record highs  as Hurricane Ivan \par(2) U.S. stocks looked to open flat on  Friday as oil prices hit another record and a mix of  technology shares failed to dispel worries about the economy.\\
    UniLM + ST\par (select) & (1) Shares of Air New Zealand Ltd., the country's largest airline, have surged after it emerged that the carrier's proposed alliance with Air New Zealand Zealand Ltd. \par (2) A group of companies led by Sony Corp of America which includes cable television's Comcast Corp. on Wednesday filed a grievance against Microsoft Corp.\\
    KEST & (1) Pfizer said it would sponsor a major clinical trial of Celebrex, its prescription arthritis pain medication, to assess the cardiovascular benefits of the drug.  \par(2) US consumer confidence rose in August following a sharp drop in the previous month, the Conference Board reported on Tuesday. \\
    \midrule
    &\textbf{Topic}: \textit{Sci/Tech}\\
    GPT2 + ST\par (select)& (1) WASHINGTON (Reuters) - The U.S. Supreme Court on  Thursday refused to consider a challenge to a law that  allows doctors to order people to put their breast implants in  public view\par(2) NEW YORK (Reuters) - U.S. stocks gained on Monday,  recovering from a day-earlier flat, as data showed U.S.-born  immigrants account for about 40 percent.\\
    UniLM + ST\par (select) & (1) Apple's new design may just be a good sign, but it's also a great day for its most popular iPod music player. The company says it shipped more than 1 million songs. \par (2) Reuters - More than 20, 000 people have been displaced since the end of August by violence in an area of North Darfur state in Sudan, a U. N. report said on Friday.\\
    KEST & (1) The Internet Corporation for Assigned Names and Numbers ( ICANN ) has modified its proposal to include some domain names in the name of the Internet Corporation for assigned Names and numbers.  \par(2) This holiday season, Apple Computers Inc. plans to open its first European retail store later this month in the capital city of Munich.\\
    \bottomrule
 \end{tabular}
 \caption{Example text for topic-controlled generation (continued).}
 \label{tab:topic-case-continue}
\end{table*}

\end{document}